\documentclass{article}

\usepackage{microtype}
\usepackage{graphicx}
\usepackage{tabularx}
\usepackage{subcaption}
\usepackage{booktabs} % for professional tables
\usepackage{multirow} % for multirow cells in tables

\usepackage{hyperref}

\usepackage{xcolor}
\usepackage{colortbl}

\usepackage[preprint]{icml2026}

\usepackage{amsmath}
\usepackage{amssymb}
\usepackage{mathtools}
\usepackage{amsthm}

\usepackage[capitalize,noabbrev]{cleveref}

\theoremstyle{plain}

\theoremstyle{definition}

\theoremstyle{remark}

\usepackage[textsize=tiny]{todonotes}

\icmltitlerunning{Shared Nature, Unique Nurture: PRISM for Pluralistic LLMs via Dynamic Epistemic Graphs}

\usepackage[most]{tcolorbox}
\usepackage{graphicx}
\usepackage{xcolor}
\usepackage{enumitem}
\tcbuselibrary{breakable}
\usepackage{float} 

\tcbset{
  novex/.style={
    colback=gray!3,
    colframe=gray!40,
    boxrule=0.6pt,
    arc=3pt,
    left=6pt,
    right=6pt,
    top=6pt,
    bottom=6pt
  }
}
\newcommand{\icon}[1]{\raisebox{-0.15em}{\includegraphics[height=1.05em]{#1}}}

\newcommand{\prismhdr}[1]{\icon{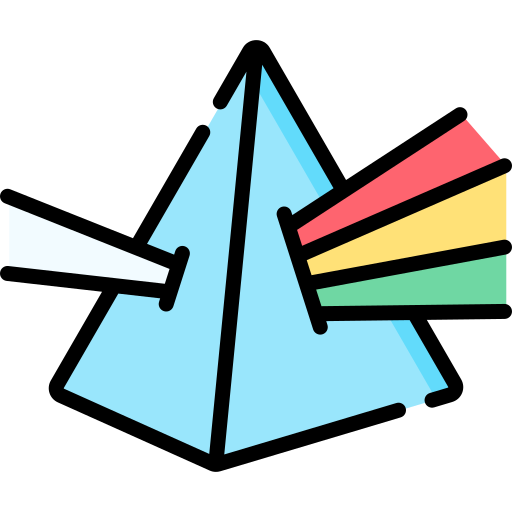}\ \textbf{PRISM (Qwen3-4B-Instruct, Distinct@10 = #1)}}
\newcommand{\qwenhdr}[1]{\icon{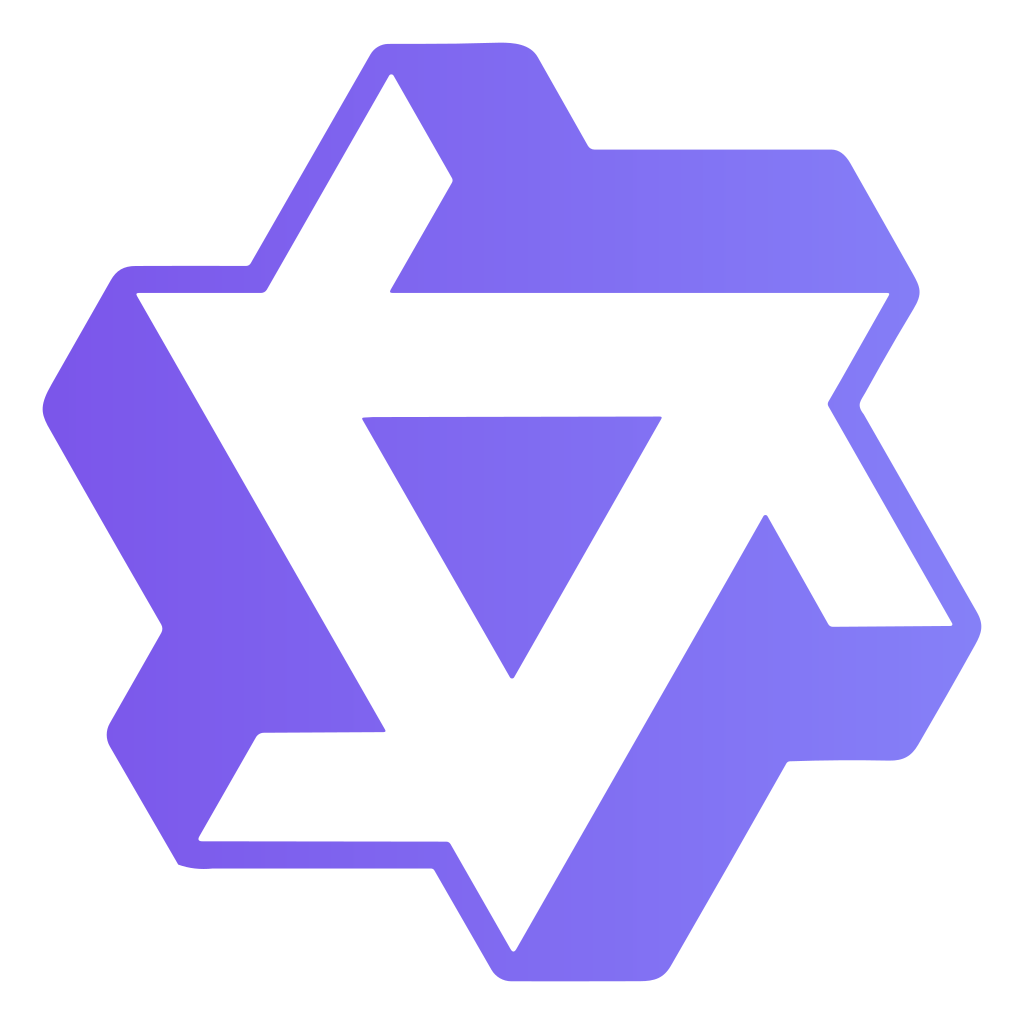}\ \textbf{Base Qwen3-4B-Instruct (Distinct@10 = #1)}}

\newcommand*\inlineimage[1]{\raisebox{-0.15\baselineskip}{\includegraphics[height=1\baselineskip]{#1}$\,$}}
\newcommand{\titleemoji}{\inlineimage{Fig/icons/prism.png}}

\begin{document}

\twocolumn[
  % \icmltitle{One World, Many Minds: \\ Inducing Pluralistic AI via Dynamic Epistemic Contexts}
  % \icmltitle{Shared Nature, Unique Nurture: \\ Individuating LLMs with Dynamic Epistemic Graphs}
  \icmltitle{\titleemoji Shared Nature, Unique Nurture: \\ PRISM for Pluralistic Reasoning via In-context Structure Modeling}



  \icmlsetsymbol{equal}{*}

  \begin{icmlauthorlist}
    \icmlauthor{Guancheng Tu}{equal,upenn}
    \icmlauthor{Shiyang Zhang}{equal,yale}
    \icmlauthor{Tianyu Zhang}{mila,udem}
    \icmlauthor{Yi Zhang}{ucsc}
    \icmlauthor{Diji Yang}{ucsc}
  \end{icmlauthorlist}

  \icmlaffiliation{upenn}{University of Pennsylvania}
  \icmlaffiliation{yale}{Yale University}
  \icmlaffiliation{ucsc}{University of California Santa Cruz}
  \icmlaffiliation{mila}{MILA}
  \icmlaffiliation{udem}{Université de Montréal}

  \icmlcorrespondingauthor{Diji Yang}{dyang39@ucsc.edu}

  \begin{center}
    Project Site: \texttt{\url{https://prism4research.com}}
  \end{center}

  \icmlkeywords{Machine Learning, ICML}

  \vskip 0.3in
]



\printAffiliationsAndNotice{}  % no special notice (required even if empty)

\begin{abstract}
% --------------------------------------------------
% Problem:
% - Creative research and open-ended generation as a core LLM capability
% - Limitations of current models in novelty and diversity
%
% Method:
% - Introduce the Creative Research System (model-agnostic, plug-and-play)
%
% Results:
% - Strong gains on NoveltyBench, IdeaBench, and Artificial Hivemind
%
% Takeaway:
% - Improves distributional creativity rather than isolated outputs
% --------------------------------------------------
Large Language Models (LLMs) are converging towards a singular Artificial Hivemind, where shared \textit{Nature} (pre-training priors) result in a profound collapse of distributional diversity, limiting the distinct perspectives necessary for creative exploration and scientific discovery. 
To address this, we propose to equip models with inference-time \textit{Nurture} (individualized epistemic trajectories) using \textbf{Epistemic Evolution} paradigm, progressing through explore, internalize, and express. 
We instantiate this via \textbf{PRISM} (\underline{P}luralistic \underline{R}easoning via \underline{I}n-context \underline{S}tructure \underline{M}odeling), a model-agnostic system that augments LLM with dynamic On-the-fly Epistemic Graphs. 
On three creativity benchmarks, PRISM achieves state-of-the-art novelty and significantly expands distributional diversity. Moreover, we evaluate the real-world utility via a challenging rare-disease diagnosis benchmark. Results demonstrate that PRISM successfully uncovers correct long-tail diagnoses that standard LLM miss, confirming that its divergence stems from meaningful exploration rather than incoherent noise.
Overall, this work establishes a new paradigm for Pluralistic AI, moving beyond monolithic consensus toward a diverse ecosystem of unique cognitive individuals capable of collective, multi-perspective discovery.
\end{abstract}

% ==================================================
\section{Introduction}
\label{sec:intro}

% The Crisis - Scaling leads to Homogeneity (The Problem)
Large Language Models (LLMs) have achieved impressive success as universal repositories of world knowledge~\cite{comanici2025gemini,singh2025openai,wang2024mmlu}. Driven by similar data and training receipt, models from diverse organizations are converging towards a highly capable intelligence in terms of knowledge exploitation~\cite{huang2025winning}. However, this convergence comes with a critical systemic risk: the emergence of an Artificial Hivemind~\citep{jiang2025artificial}. 
Following the established alignment techniques (e.g., Supervised-Finetuning, RLHF~\cite{ouyang2022training}, etc), under the intense convergence pressure to the human-defined objectives, models aggressively collapse onto a narrow band of safe reasoning patterns, stripping them of any potential for individuality~\cite{zhang2025noveltybench,yang2025many,yue2025does,kirk2023understanding}.
% As models approach the mean of their vast training corpora, their responses to open-ended inquiries exhibit a profound distributional collapse~\cite{kirk2023understanding}. 
% As a result, LLMs' responses to open-ended inquiries exhibit a profound distributional collapse~\cite{kirk2023understanding}
Altering this behavior at the training stage is fraught with challenges: optimizing for the ambiguous metric of ``creativity'' risks destabilizing the delicate balance between exploration and exploitation~\cite{franceschelli2025creativity}, model training comes with cost, and may potentially breaks the well-tuned behaviors such as safety alignment. Consequently, when every LLM thinks homogeneously, the boundaries of AI-augmented creative exploration and scientific discovery are locked.

\begin{figure}[t]
  \centering
  \includegraphics[width=0.85\columnwidth]{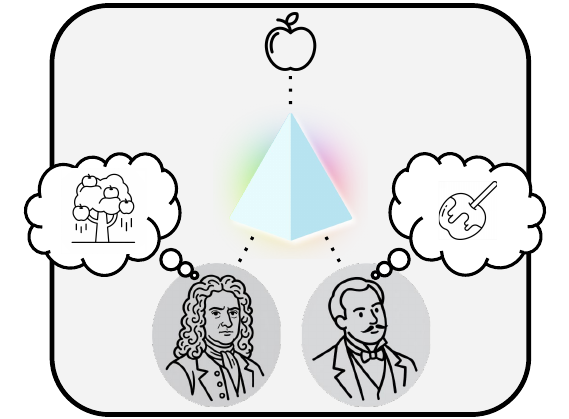}
  \caption{Divergent cognitive responses to a common stimulus: Newton's gravity vs. Kolb's candy apple, both inspired by an apple. Analogous to an optical prism, our PRISM framework aims to restore this pluralism by refracting shared pre-training knowledge into distinct, individualized epistemic trajectories.}
  \label{fig:teaser}
  \vspace{-2.3pt}
\end{figure}

% The Root Cause - Nature (Shared) vs. Nurture (Individual)
In human cognition, shared world knowledge does not preclude individual distinctiveness. As illustrated in Figure~\ref{fig:teaser}, consider the divergence between scientist Isaac Newton and candy maker William Kolb observing the same apple fall. While they share the same physical reality, Newton perceives the laws of gravity, whereas Kolb perceives the receipt of candy apple. This divergence is not due to biological differences, but rather from their distinct epistemic contexts and accumulated life experiences. Current LLMs, however, act with inter-model homogeneity. They possess encyclopedic world knowledge but lack the distinct, individual trajectories that shape unique perspectives. 

% The Solution - Simulating Cognitive Trajectories (The Methodology)
Motivated by the pluralistic intelligence of humans, we propose the \textbf{Epistemic Evolution} paradigm. Distinguishing between AI models derived from \textit{nature} (shared pre-training) and \textit{nurture} (individual experience), this paradigm simulates a localized cognitive lifecycle, cultivating a unique experience to shape the model's perspective\footnote{More discussion from a cognitive dynamics perspective is available in Appendix~\ref{appendix:pluralish}.}.
We instantiate this via PRISM (Pluralistic Reasoning via In-context Structure Modeling), a framework that intervenes at inference to construct the unique trajectory.
Starting from individualized epistemic seeds, the system performs a ``wild search'' to acquire heterogeneous information.
Rather than simply concatenating these findings, PRISM iteratively updates them into an on-the-fly epistemic graph~\cite{nikooroo2025belief}.
During the generation phase, the topological structure forces the model to traverse explicit reasoning paths, connecting distant concepts; thereby establishing a distinct, internally consistent perspective.

Empirically, we demonstrate that this structured \textit{nurture} effectively counteracts model collapse. Across the Artificial Hivemind~\cite{jiang2025artificial}, NoveltyBench~\cite{zhang2025noveltybench}, and IdeaBench~\cite{guo2025ideabench} benchmarks, PRISM diversifies response distributions and achieves state-of-the-art novelty scores. Moreover, on RareBench~\cite{chen2024rarebench}, we show that this divergence is not unconstrained gibberish but meaningful exploration; by enforcing graph-mediated reasoning, PRISM succeeds in identifying long-tail diagnostic paths that base LLM misses.

We summarize our key contributions as follows: (1) We propose Epistemic Evolution, a novel paradigm that equips LLMs with unique cognitive trajectories to break the Artificial Hivemind; (2) We introduce PRISM % \footnote{All PRISM components as illustrated in Figure~\ref{fig:pipeline} will be open-sourced to facilitate future research.}
, a model-agnostic framework that individualizes inference via On-the-fly Epistemic Graphs; (3) PRISM achieves SOTA performance across creativity, scientific discovery, and diagnosis benchmarks, establishing a foundational framework for Pluralistic AI in machine-assisted discovery.

\section{PRISM for Pluralistic LLMs}
\label{sec:prism}

% We instantiate the theoretical framework defined in Section~\ref{sec:theory} into a concrete, model-agnostic engineering architecture: \textbf{PRISM} (\textbf{P}luralistic \textbf{R}easoning via \textbf{I}n-context \textbf{S}tructure \textbf{M}odeling).

% PRISM is a ready-to-use system that operationalizes the "Nature vs. Nurture" paradigm through a two-stage cognitive pipeline: \emph{Cognitive Explosion} (to break the static prior) followed by \emph{Epistemic Structuring} (to organize the dynamic context). Formally, the system outputs a final response $y$ conditioned on an intermediate graph $G$:
% \[
% y = f_{\text{LLM}}(q, G)
% \]
Our idea is grounded in a straightforward principle: distinct intelligence emerges not just from \textit{nature}, but also from \textit{nurture}. 
% Current LLMs converge towards an Artificial Hivemind because they rely almost exclusively on the static, averaged priors crystallized during pre-training. To break this symmetry without the prohibitive cost of retraining, we posit that we must intervene at the inference stage. Instead of forcing the model to rely solely on its internal parameters, we simulate a unique "life experience" for every query. By dynamically constructing an epistemic trajectory (i.e., a structured journey of exploration that mimics how a human gathers and connects information), we can shift the model’s focus from the most probable, consensus-based answer to a diverse, contextually grounded perspective. 
Formally, we define the Epistemic Evolution paradigm, which unfolds in three abstract phases analogous the process of intellectual individuation (or human sensemaking). 
\begin{itemize} 
\item Phase I: Experiencing (Exploration). Just as individuals are shaped by exposure to distinct environments, the model is supposed to acquire heterogeneous information. This phase prioritizes dispersion over relevance, simulating the stochastic nature of life experiences.
\item Phase II: Cognitive Internalization (Exploitation). Raw experience is merely noise until processed. In this phase, the system must organize scattered observations into a stable mental state, and transform transient data into a structured cognitive context. 
\item Phase III: Contextualized Expression (Generation). Finally, the model articulates its response conditioned on this unique mental state. The output is no longer a retrieval from memory, but a synthesis derived from the constructed perspective. 
\end{itemize} 
This paradigm shifts the locus of individuation from static weights to dynamic context: the implementation method at each stage can evolve, but we expect the principle of unique nurture to remain constant. By simulating the human cognitive cycle, where exploration undergoes experiential internalization into a structured belief system, the model transcends passive retrieval to forge its own ``lived'' trajectory, ultimately manifesting as an idiosyncratic perspective.

\begin{figure*}[t]
    \centering
    \includegraphics[width=0.80\textwidth]{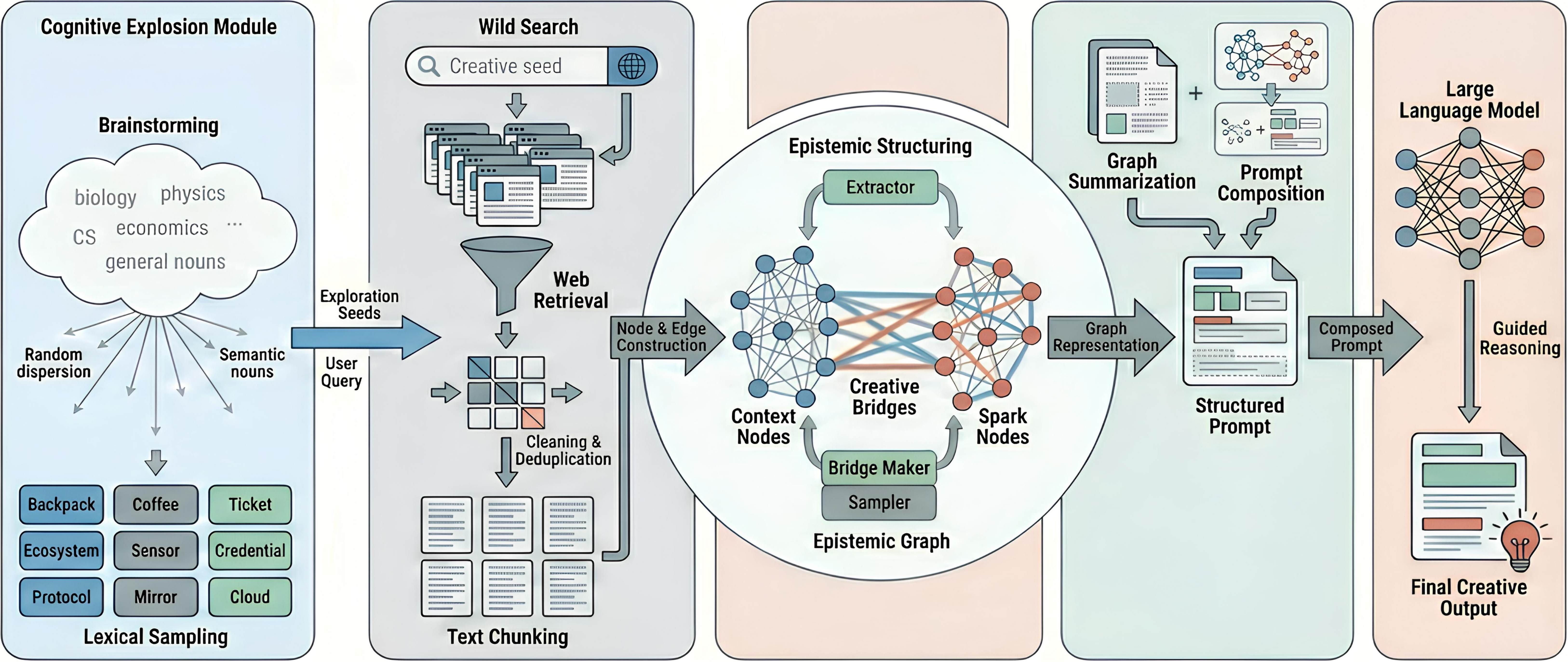}
    \caption{The PRISM Pipeline. The system first performs \textbf{Cognitive Explosion} via wild search to break local relevance, followed by \textbf{Epistemic Structuring} where Context Nodes and Spark Nodes are bridged via cognitive operators (Mapping, Blending, Inversion).}
    \label{fig:pipeline}
\end{figure*}

We operationalize this principle into \textbf{PRISM} (Pluralistic Reasoning via In-context Structure Modeling), a ready-to-use model-agnostic framework as illustrated in Figure~\ref{fig:pipeline}. 
In the remainder of this section, we present the technical details of its three major components, as well as the rationale behind the design. First, we introduce the Cognitive Explosion module (Section~\ref{sec:wildsearch}), which employs high-entropy sampling to break the local minima of the model’s pre-training priors. Next, we describe Epistemic Structuring (Section~\ref{sec:graph_construction}), where raw information is stabilized into a coherent Knowledge Graph via cognitive operators. Finally, we cover the Conditional Generation (Section~\ref{sec:synthesis}), explaining how the graph acts as a topological constraint to steer the final generation.

\subsection{Phase I: Cognitive Explosion (Breaking Nature)}
\label{sec:wildsearch}

To induce the \textit{Nurture} component, we must first liberate the model from the local minima of its pre-training prior (\textit{Nature}). We achieve this through a \textbf{Cognitive Explosion Module}, instantiated as a \emph{Wild Search} mechanism. Unlike standard retrieval, which optimizes for relevance, this module optimizes for semantic dispersion.

\noindent \textbf{Stochastic Lexical Sampling.}
To inject high-entropy priors, we adopt a randomized seeding strategy that samples a small set of lexical units from a global noun vocabulary. Specifically, we randomly sample $k \in \{3\}$ nouns to serve as stochastic perturbations of the original query context. This design intentionally introduces uncontrolled semantic variation, simulating diverse cognitive starting points induced by different knowledge backgrounds and personal experiences. Each sampled seed set $S$ thus represents a distinct epistemic initialization, analogous to individualized developmental trajectories.

\noindent \textbf{Wild Retrieval \& Filtering.}
Each sampled seed is independently issued as a query to a large-scale search platform, initiating parallel exploratory retrieval processes. This exposes the system to heterogeneous and weakly correlated information sources beyond the immediate semantic neighborhood of the original query. To ensure quality, the retrieved documents undergo a strict cleaning pipeline. URL-level and hash-based deduplication are applied first, followed by content filtering to remove navigational, commercial, and low-entropy pages. Finally, the remaining documents are segmented into overlapping chunks using a sliding window to form the candidate corpus $C$.

\subsection{Phase II: Epistemic Graph Construction (Structuring Nurture)}
\label{sec:graph_construction}

Raw divergence produces noise. To transform this noise into a structured \textit{Nurture} context, we construct the Epistemic Graph $G = (V_c, V_s, E)$ through a multi-stage extraction and bridging procedure. This process acts as a critical \textbf{Epistemic Stabilization} mechanism, constraining exploration within interpretable conceptual structures to maintain coherence while traversing distant semantic regions.

\noindent \textbf{Node Extraction: Anchors and Sparks.}
The graph topology is defined by two node classes. \textbf{Context Nodes ($V_c$)} are extracted from the user query $q$ to represent immutable constraints and core entities. \textbf{Spark Nodes ($V_s$)} are extracted from the retrieved corpus $C$ using a specialized prompt that identifies operational mechanisms (``how things work"), salient properties, and emergent byproducts:
\[
V_s \leftarrow \bigcup_{c \in C} \textsc{SparkExtractor}(c, q)
\]
These nodes represent external \textit{Nurture} signals—novel mechanisms that the base model may not have prioritized.

\noindent \textbf{Creative Edge Generation via Cognitive Operators.}
Edges $E$ are constructed not by co-occurrence, but by specialized \textbf{Cognitive Operators} that simulate analogical reasoning. \textbf{Mapping ($\xrightarrow{M}$)} transfers mechanisms across domains (e.g., mapping a biological ``viral spread" to a marketing problem). \textbf{Blending ($\xrightarrow{B}$)} combines attributes from a Context Node and a Spark Node into a novel composite. \textbf{Inversion ($\xrightarrow{I}$)} introduces productive tension by identifying Spark Nodes that functionally oppose a Context Node.

\noindent \textbf{Sampling and Topological Constraints.}
To avoid combinatorial explosion and semantic collapse, we enforce strict topological constraints during edge generation. Specifically, we prohibit context--context ($V_c \leftrightarrow V_c$) connections, as these represent the static problem definition. Instead, we prioritize heterogeneous pairs ($V_c \leftrightarrow V_s$) and spark--spark ($V_s \leftrightarrow V_s$) interactions. Beyond regularization, these structured connections actively promote creative collisions between weakly related nodes, inducing novel contextual compositions and expanding the effective diversity of the graph. This design prevents degeneration into trivial semantic loops and ensures that the \textit{Nurture} component continuously reshapes the interpretation of the \textit{Nature} component.

\paragraph{Design Rationale: Why Graph Topology?}  
Our choice of a graph structure serves as an explicit \textbf{reasoning substrate} rather than a black-box retrieval dump. Without this structural digestion, direct injection of raw retrieved terms risks ``semantic normalization,'' where the LLM's robust pre-training bias treats high-entropy novelty as noise to be corrected---analogous to how models auto-correct typos---thereby reverting the output to the mean. By enforcing graph connectivity, we implement a computational human prior: the graph allows the system to genuinely \textbf{digest} the raw retrieval noise, actively forming associations and establishing connections. This process effectively \textbf{internalizes} external signals into the LLM's own ``background experiences,'' thereby fostering a \textbf{unique, synthesized perspective} that preserves novelty against the model's tendency to normalize.

\subsection{Phase III: Conditional Generation}
\label{sec:synthesis}

The final stage is the utilization of the constructed graph to condition the base model.

\noindent \textbf{Graph Serialization and Inference.}
The graph $G$ is serialized into a textual representation $\hat{G}$ that explicitly exposes the bridging logic. The base model performs inference on this augmented context, $y \leftarrow M(q, \hat{G})$. This externalizes intermediate creative reasoning into an explicit symbolic structure, facilitating controllable synthesis and interpretability.

\section{Experiments: Creativity and Discovery}
\label{sec:main_experiments}

% We evaluate the effectiveness of our PRISM System in promoting creative exploration and mitigating distributional collapse across three complementary dimensions: distributional diversity (\textit{Artificial Hivemind}), open-ended creativity (\textit{NoveltyBench}), and scientific discovery (\textit{IdeaBench}). Our evaluation aims to examine not only whether our system improves metric scores, but also whether it induces systematic, structured exploration of the semantic space without compromising utility.

We evaluate the effectiveness of the PRISM system in fostering creative exploration and mitigating distributional collapse across three complementary dimensions: distributional diversity (\textit{Artificial Hivemind}), open-ended creativity (\textit{NoveltyBench}), and scientific discovery (\textit{IdeaBench}). Beyond merely improving benchmark scores, we examine whether the system can internalize exploration into a cohesive experience, ensuring that its actions are refracted through a distinctive perspective without drifting from the task's logical constraints; we provide more qualitative analysis of this phenomenon in Appendix~\ref{app:novelbench_examples}.

% --------------------------------------------------
\subsection{Experimental Setup}
\label{sec:exp_setup}

\noindent \textbf{Base Models.} 
To ensure the generality of our framework, we employ a diverse set of language models spanning different architectures, training paradigms, and scales. Our evaluation suite includes both proprietary models and open-weight variants to verify that our method is not dependent on specific model weights or hidden prompts. Refer to Appendix~\ref{app:models} for the list of model versions and access details.

\noindent \textbf{Protocol.} 
% Across all experiments, model weights remain fixed. We compare vanilla generation (direct prompting with high temperature decoding) against generation augmented by our system. 
To ensure a fair comparison, all decoding hyperparameters (e.g., temperature, max tokens, etc) are strictly aligned with the original report or evaluation codebase of each benchmark. This setup ensures that performance gains are attributable solely to PRISM. Detailed configurations are provided in Appendix~\ref{sec:appendix_config} for reproducibility.

% --------------------------------------------------
\noindent \subsection{Distributional Diversity: Artificial Hivemind}
\label{sec:hivemind}

\noindent \textbf{Benchmark Overview.}
Artificial Hivemind characterizes the collective behavior of LLMs, specifically measuring the tendency of models to converge on a narrow set of safe responses. We select 15 representative open-ended questions and generate 50 distinct responses per question across four base models to map the distributional landscape.

\begin{figure}[!t]
    \centering
    \includegraphics[width=0.9\columnwidth]{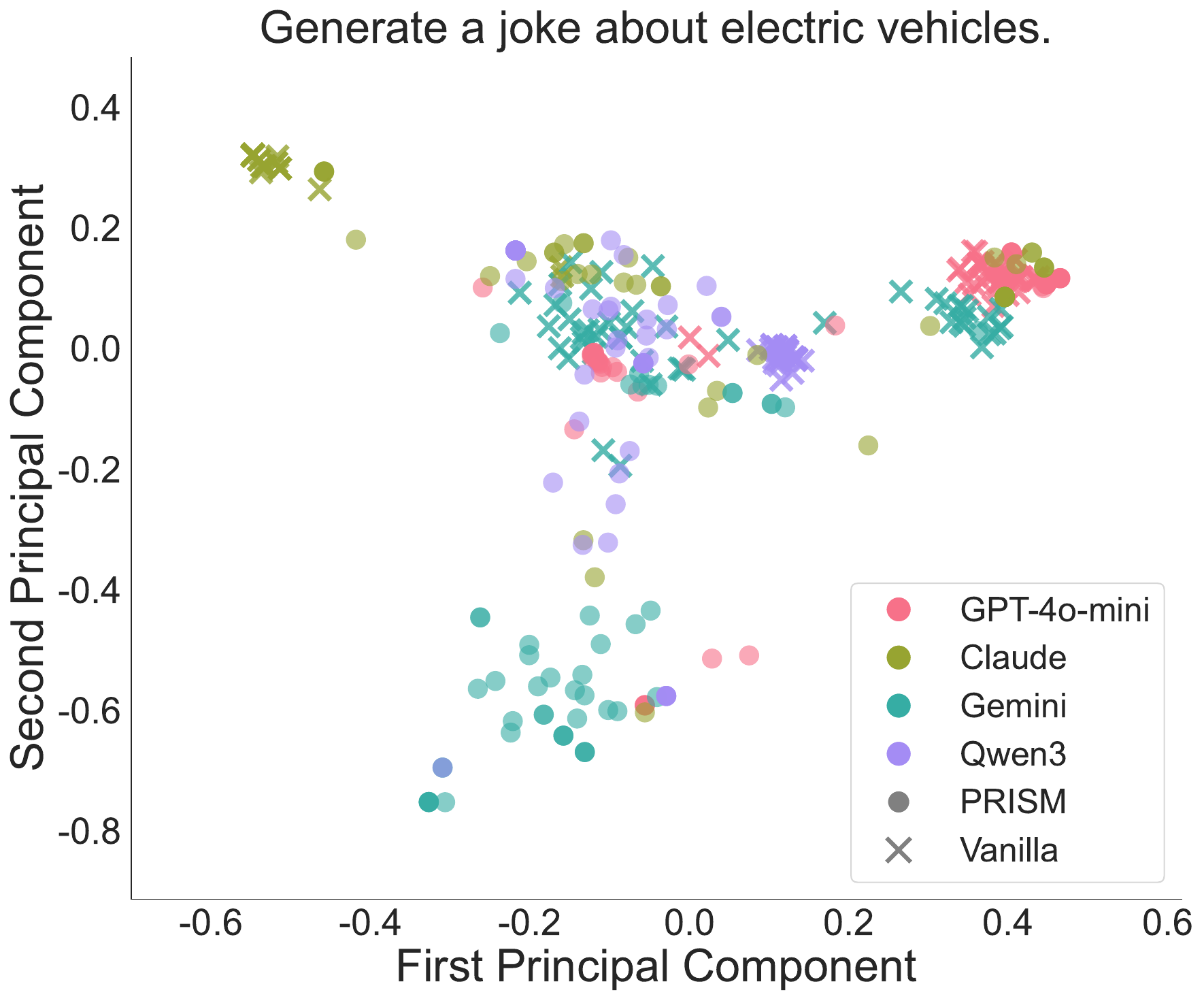}
    \caption{PCA visualization of response distributions. PRISM (in dot) produces multi-centered, elongated distributions compared to the concentrated clusters of vanilla generation.}
    \label{fig:hivemind_pca}
\end{figure}

\noindent \textbf{Analysis and Results.}
We analyze the semantic spread of outputs using Principal Component Analysis (PCA) on sentence embeddings. As illustrated in Figure~\ref{fig:hivemind_pca}, vanilla generations tend to concentrate tightly around a small number of dominant semantic modes, reflecting the hivemind collapse. In contrast, our system produces multi-centered and elongated distributions that cover substantially broader regions of the embedding space.

Quantitatively, we observe a significant reduction in \textbf{Intra-Model Similarity}, which indicates less self-repetition within a single model's outputs. Furthermore, the \textbf{Inter-Model Similarity} decreases, suggesting reduced homogenization between different model families. These findings demonstrate that our framework induces structured semantic exploration driven by individualized epistemic trajectories rather than superficial stochasticity.

\noindent \textbf{Qualitative Distribution Analysis.}
Figure~\ref{fig:hivemind_pca} illustrates the visual impact of PRISM on semantic exploration.
In the vanilla setting, Claude, Qwen, and \texttt{gpt-4o-mini} exhibit severe mode collapse, with 50 responses per prompt concentrating into nearly singular, high-density clusters.
While vanilla Gemini displays higher baseline variance, it still shows noticeable aggregation in the upper-right quadrant.

Applying PRISM triggers a dramatic transition from these static templates to expansive, multi-centered distributions.
This shift is most evident for the highly concentrated models (Claude and Qwen), which begin to navigate distinct semantic trajectories across the embedding space.
Even for Gemini, PRISM visibly smooths and extends its reach, producing more uniform and elongated coverage.

These results confirm that our framework effectively mitigates the ``Artificial Hivemind'' effect, compelling models to explore diverse conceptual directions that vanilla generation typically ignores. Detailed qualitative examples and additional PCA visualizations across more prompts are provided in Appendix~\ref{app:novelbench_examples} and Appendix~\ref{app:hivemind_more_figs}.

\begin{figure}[t]
    \centering
    \includegraphics[width=\columnwidth]{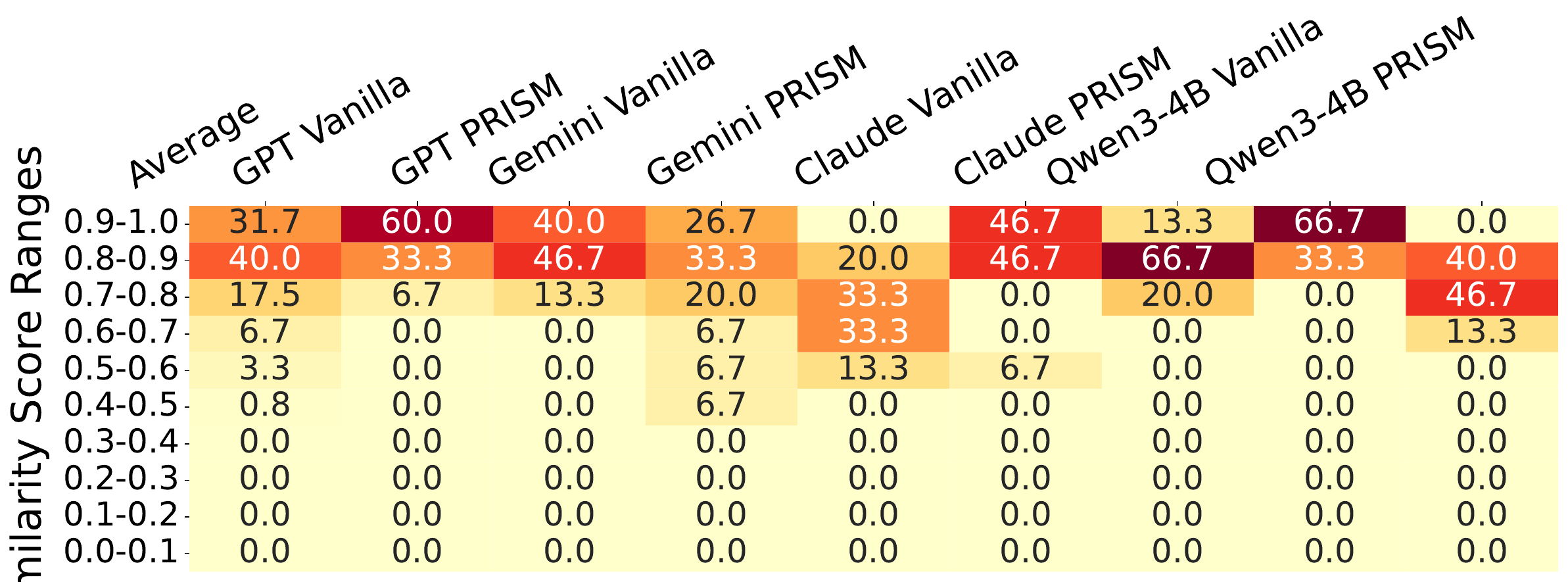}
    \caption{Intra-model diversity heatmap on Artificial Hivemind.}
    \label{fig:intra}
    \vspace{-3pt}
\end{figure}

\noindent \textbf{Intra-Model Similarity Analysis.}
To quantitatively assess self-repetition within individual models, we compute the pairwise cosine similarity among all responses generated by the same model and visualize the results using similarity-range histograms (Figure~\ref{fig:intra}).

Across all model families, vanilla generation exhibits strong concentration in high-similarity intervals (0.8--1.0), indicating frequent semantic redundancy.
This confirms that baseline models tend to recycle highly similar response patterns even under stochastic sampling.

After applying our system, the mass of similarity scores shifts consistently toward lower ranges.
High-similarity intervals are significantly reduced, while medium- and low-similarity regions become more prominent.
This demonstrates that our method effectively suppresses self-reinforcing generation loops and promotes internal diversity within each model.

Importantly, this improvement is observed across proprietary and open-weight models, suggesting that our approach generalizes beyond specific architectures or training paradigms.

\begin{figure}[t]
    \centering
    \includegraphics[width=0.8\columnwidth]{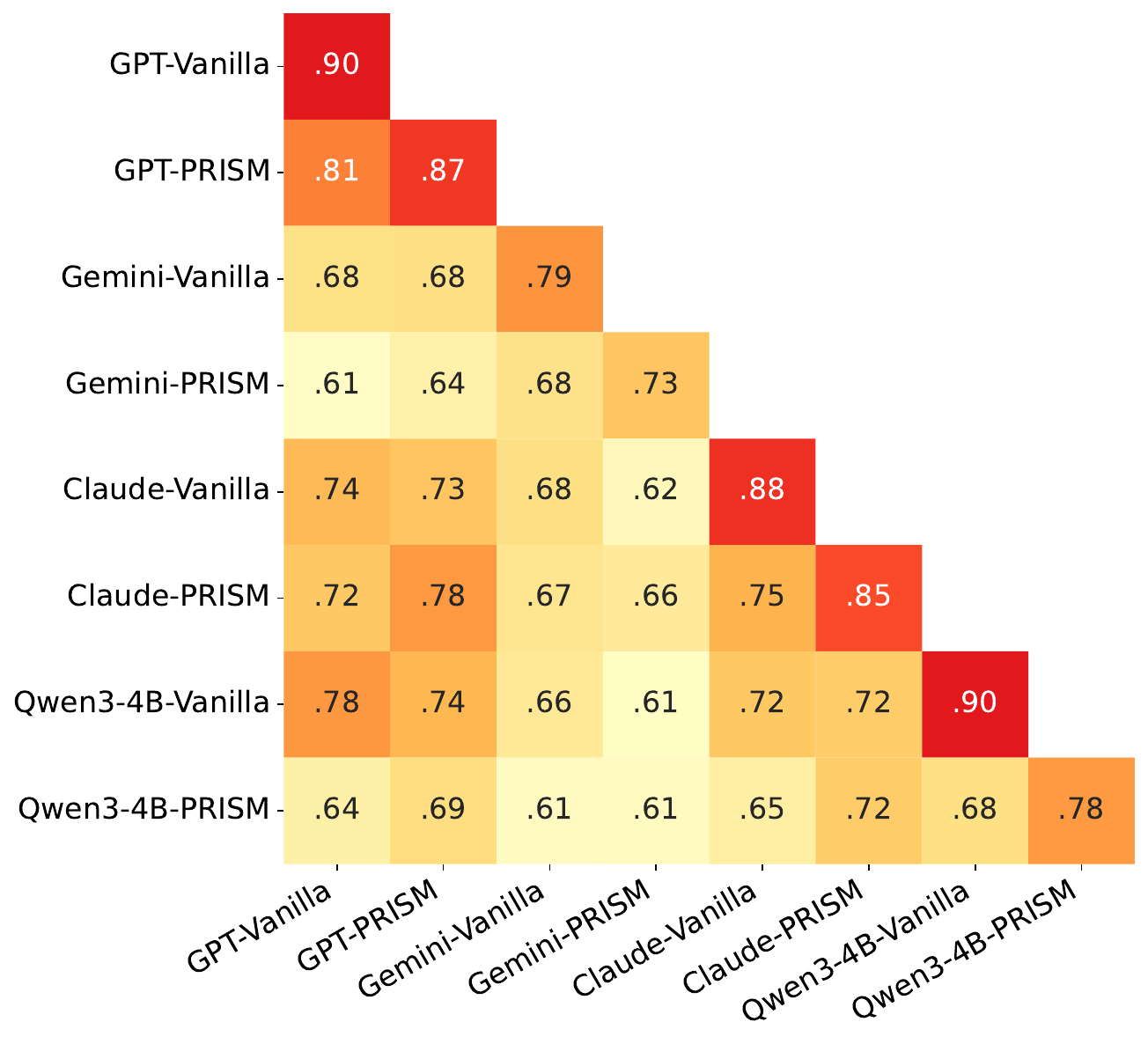}
    \caption{Inter-model similarity heatmap on Artificial Hivemind.}
    \label{fig:inter}
    \vspace*{-3pt}
\end{figure}

\noindent \textbf{Inter-Model Similarity Analysis.} 
To assess cross-model homogenization, we measure pairwise similarities between responses across different model families (Figure~\ref{fig:inter}). 

Under Vanilla settings, we observe high similarity (often $>0.75$) even between distinct models, reflecting the Artificial Hivemind effect where shared pre-training and alignment objectives lead to convergent, surface-level outputs. Our framework consistently reduces these similarities, creating a more diffuse matrix and lower off-diagonal entries. 

A notable insight arises from comparing \textit{within-model} shifts against \textit{cross-model} baselines: for instance, the similarity between Qwen3-4B-Vanilla and Qwen3-4B-PRISM (0.68) is significantly lower than that between Qwen3-4B-Vanilla and GPT-Vanilla (0.78). This suggests that our system induces a unique perspective that creates greater divergence than the inherent architectural or data differences between major model families. By externalizing and evolving unique epistemic trajectories, the framework effectively facilitates structured semantic exploration rather than reproducing globally dominant patterns.

\begin{table}[!t]
\centering
\caption{Result on NoveltyBench (Distinct score) and IdeaBench (Novelty score).}
\resizebox{0.8\columnwidth}{!}{
    \begin{tabular}{lcc}
        \toprule
         &
        NoveltyBench &
        IdeaBench \\
        \midrule
        
        Qwen3-4B-Instruct (Vanilla) & 3.09 & 0.72 \\ 
        \rowcolor[rgb]{ .867, .922, .969}~~~~$\hookrightarrow$~PRISM & \textbf{4.48} & \textbf{0.96} \\ 
        CrPO-sft-LLaMA-3.1 & 7.35 & - \\
        \rowcolor[rgb]{ .867, .922, .969}~~~~$\hookrightarrow$~PRISM & \textbf{7.67} & - \\ 
        GPT-4o-mini (Vanilla) & 2.65 & 0.45 \\ 
        \rowcolor[rgb]{ .867, .922, .969}~~~~$\hookrightarrow$~PRISM & \textbf{3.41} & \textbf{0.65} \\ 
        \bottomrule
    \end{tabular}
}
\label{table:main_results}
\end{table}

% --------------------------------------------------
\subsection{Open-Ended Creativity: NoveltyBench}
\label{sec:NoveltyBench}

\noindent \textbf{Benchmark Overview.}
NoveltyBench provides a standardized evaluation for open-ended QA using the \textbf{Distinct-$k$} metric, which quantifies diversity by measuring the number of unique responses generated within a set of $k$ candidates. For more detailed documentation, see Appendix \ref{sec:appendix_config}.

\noindent \textbf{Results.} 
Table~\ref{table:main_results} summarizes the performance across representative models, including CRPO \cite{ismayilzada2025creative}, a post-training baseline for diversity-oriented optimization that currently stands as the SOTA on NoveltyBench. We observe consistent improvements in the Distinct score across all evaluated models. Notably, relative gains are more pronounced for smaller and mid-scale models; for instance, \texttt{gpt-4o-mini} achieves a 28\% increase in its Distinct score, indicating that dynamic epistemic conditioning can effectively compensate for limited intrinsic diversity. Furthermore, the performance trend demonstrates that PRISM can seamlessly wrap any model to further amplify their response diversity.

% --------------------------------------------------
\subsection{Scientific Discovery: IdeaBench}
\label{sec:ideabench}

\noindent \textbf{Benchmark Overview.}
To test creativity in a rigorous domain, we employ \textit{IdeaBench}, which evaluates the novelty of generated research hypotheses conditioned on academic literature. Unlike open-ended QA, this task requires the model to synthesize new ideas based on a target paper's title and its cited abstracts.

\noindent \textbf{Metrics.} 
We report the \textit{Novelty Insight Score} (NIS) to evaluate the distinctiveness of the generated hypotheses. For each instance, three candidate ideas are generated and ranked alongside the original idea from the target paper as a benchmark by an LLM-based evaluator. These ordinal rankings are then transformed into a final quantitative score, reflecting the relative novelty of the generated content against existing literature.

\noindent \textbf{Results.} As shown in Table~\ref{table:main_results}, wrapping vanilla models within the PRISM framework leads to a universal and marked boost in \textit{Novelty Insight Scores}. Specifically, \texttt{gpt-4o-mini} achieves a 44.4\% improvement, illustrating that the epistemic graph empowers models to break away from consensus-heavy outputs. This confirms PRISM's ability to foster ``lived experiences" that elicit unique research perspectives, shifting the model from pattern replication toward the generation of truly novel scientific ideas. Detailed metrics across all baselines are provided in Table~\ref{tab:combined_results}.

% ==================================================
% ==================================================
\section{Real-world Application: Rare Disease Diagnosis}
\label{sec:rarebench}

While our previous experiments demonstrate PRISM's capacity for divergent creativity in open-ended domains, real-world deployment requires that such exploration remain strictly grounded in evidence. We move from the expansive exploration of \textit{NoveltyBench} to the RAMEDIS subset of RareBench \citep{chen2024rarebench}. This represents a rigorous test of the Epistemic Evolution paradigm (Section~\ref{sec:prism}), as accurate rare disease diagnosis requires traversing sparse, long-tail knowledge regions where parametric ``Nature'' (pre-training priors) often fails.

\subsection{The Dilemma of Conservative Models}
In clinical settings, standard LLMs (Vanilla Models) exhibit a \textit{conservative bias}, fixating on common conditions to minimize immediate error, a behavioral manifestation of the Artificial Hivemind discussed in Section~\ref{sec:intro}. While this strategy yields a low Mean Rank for frequent diseases, it results in catastrophic recall for the rare cases that constitute the ``long tail.'' To diagnose effectively, the agent must break this conservative mold by simulating a specialized \textit{individualized epistemic trajectory}, expanding its candidate space to ensure rare diagnostic paths are not prematurely pruned.

\subsection{Methodology: Disentangling Search, Structure, and Intent}

To isolate the contributions of structural organization versus exploration quality, we first establish two control baselines: \textbf{Search-Only (Flat RAG)}, which uses syntactic seeds and basic context concatenation to test whether raw information alone can break pre-training priors. We then evaluate \textbf{PRISM (Syntactic)}, retaining the same seeds but organizing retrieved data via the epistemic graph; this isolates the specific role of the \textit{Cognitive Internalization} phase in stabilizing external noise. Finally, the \textbf{PRISM (Expert)} framework upgrades the \textit{Experiencing} phase by deploying a virtual consultation panel (detailed in Appendix~\ref{sec:rareexp_setup}) of LLM-simulated specialists. These experts replace static keywords with n-domain semantic seeds, actively steering the Cognitive Explosion toward long-tail diagnostic regions and grounding the epistemic graph in deep medical reasoning (see Appendix~\ref{app:rare_examples}).

\subsection{Results: Breaking the Conservative Trap}
Table~\ref{tab:rarebench_results} presents the performance metrics. The divergence between these paradigms validates the structural design of PRISM through three critical insights.

\textbf{The Failure of Flat Retrieval.} The \textit{Search-Only} baseline performs worse than the zero-shot model on Recall@1 (14.0\% vs. 16.0\%). This indicates that unstructured retrieval without the stabilizing effect of an epistemic graph may mislead the LLM, as irrelevant documents clutter the reasoning context---a result consistent with recent findings on RAG-induced degradation \cite{zeng2025worse}.

\textbf{Graph Structure as a Precision Filter.} \textbf{PRISM (Syntactic)} repairs this degradation (16.7\% Recall@1) and boosts Recall@10 to 39.2\%. This confirms that the epistemic graph effectively performs \textit{internalization}, suppressing noise and allowing the model to benefit from long-tail evidence without being overwhelmed.

\textbf{Expert Intent and the Trade-off of Discovery.} The most striking result comes from \textbf{PRISM (Expert)}, which drives a surge in Recall@10 to 52.0\%. Unlike the Vanilla Model, which provides a ``confident but narrow'' output (Mean Rank 1.50 but low recall), PRISM prioritizes \textit{exploratory coverage}. The increase in Mean Rank (2.92) is a deliberate byproduct of this clinical trade-off: PRISM provides a comprehensive differential diagnosis list, placing the correct rare disease in the candidate set significantly more often (+20\% gain over Vanilla). In rare disease contexts, this transition from monolithic consensus to broad, grounded discovery is essential for clinical utility.

\begin{table}[t]
\centering
\caption{Diagnostic performance on RareBench (RAMEDIS). While Flat RAG degrades performance due to noise, introducing the PRISM graph structure (Syntactic) improves recall. Furthermore, equipping PRISM with \textbf{Expert Intent} (Semantic Seeds) strengthens the Epistemic Evolution framework, significantly extending the coverage of long-tail diagnoses.}
\label{tab:rarebench_results}
\vskip 0.105in
\begin{center}
\resizebox{\columnwidth}{!}{%
\begin{tabular}{llccc}
\toprule
Strategy & Mechanism & Recall@1 ($\uparrow$) & Recall@10 ($\uparrow$) & Mean Rank ($\downarrow$) \\
\midrule
Vanilla Model & Parametric Prior (Zero-shot) & 16.0\% & 32.0\% & 1.50 \\
Search-Only & Flat RAG (Syntactic Seeds) & 14.0\% & 28.0\% & 1.79 \\
PRISM (Syntactic) & Graph + Keyword Seeds & 16.7\% & 39.2\% & 2.50 \\
\rowcolor[rgb]{ .867, .922, .969} \textbf{PRISM (Expert)} & \textbf{Graph + Expert Intent} & \textbf{22.0\%} & \textbf{52.0\%} & \textbf{2.92} \\
\bottomrule
\end{tabular}}
\end{center}
\vskip -0.1in
\end{table}
% ==================================================
\section{In-depth Analysis}
\label{sec:analysis}

While Section~\ref{sec:main_experiments} demonstrated the overall effectiveness of PRISM across benchmarks, this section zooms in to dissect the system's operational mechanisms and boundary conditions. Specifically, we isolate the impact of the Epistemic Graph, model scale, the breadth of exploration seeds, and lexical diversity to provide insight into the critical design trade-offs governing the system's effectiveness.

% \begin{table}[t]
% \centering
% \caption{Comprehensive sensitivity analysis. We analyze the impact of \textbf{Model Scale} (using the Qwen3 family) and \textbf{Seed Source} (using different lexical pools) on creativity and discovery metrics.}
% \label{tab:sensitivity_combined}
% \resizebox{\columnwidth}{!}{%
% \begin{tabular}{llcc}
% \toprule
% \textbf{Analysis Dimension} & \textbf{Configuration} & \textbf{NoveltyBench} & \textbf{IdeaBench} \\
% \midrule
% \multicolumn{4}{l}{\textit{\textbf{I. Effect of Model Scale \& Paradigm}}} \\
% Qwen3-1.7B & Vanilla & 4.82 & 0.39 \\
%  & \cellcolor[rgb]{.867, .922, .969}+ PRISM & \cellcolor[rgb]{.867, .922, .969}5.17 & \cellcolor[rgb]{.867, .922, .969}0.65 \\
% \cmidrule{1-4}
% Qwen3-4B-Instruct & Vanilla & 5.05 & 0.63 \\
%  & \cellcolor[rgb]{.867, .922, .969}+ PRISM & \cellcolor[rgb]{.867, .922, .969}5.48 & \cellcolor[rgb]{.867, .922, .969}0.93 \\
% \cmidrule{1-4}
% Qwen3-4B-Thinking & Vanilla & 3.14 & 0.70 \\
%  & \cellcolor[rgb]{.867, .922, .969}+ PRISM & \cellcolor[rgb]{.867, .922, .969}\textbf{6.61} & \cellcolor[rgb]{.867, .922, .969}\textbf{0.85} \\

% \midrule
% \midrule

% \multicolumn{4}{l}{\textit{\textbf{II. Effect of Lexical Seed Source}}} \\
% Lexical Pool & General Pool (Standard) & 4.42 & 0.57 \\
%  & Multi-Domain Pool & 4.41 & 0.61 \\
% \bottomrule
% \end{tabular}
% }
% \end{table}

\begin{table}[h]
\centering
\caption{Ablation Study: Contribution of the Epistemic Graph structure vs. Flat RAG.}
\label{tab:ablation_core}
\resizebox{0.9\columnwidth}{!}{%
\begin{tabular}{lccc}
\toprule
Method & Benchmark & Novelty & Improvement \\
\midrule
Search-Only (Flat RAG) & IdeaBench & 0.49 & -- \\
\rowcolor[rgb]{ .867, .922, .969} \textbf{PRISM (Full Graph)} & IdeaBench & \textbf{0.57} & \textbf{+16.33\%} \\

Search-Only (Flat RAG) & NoveltyBench & 4.38 & -- \\
\rowcolor[rgb]{ .867, .922, .969} \textbf{PRISM (Full Graph)} & NoveltyBench & \textbf{4.42} & \textbf{+0.91\%} \\

\bottomrule
\end{tabular}}
\end{table}

\begin{table}[t]
\centering
\caption{System Characterization and Design Implications. We analyze the synergistic impact of \textbf{Model Scale}, \textbf{Seed Count}, and \textbf{Lexical Seed Source} on creativity and discovery metrics.}
\label{tab:sensitivity_combined}
\resizebox{\columnwidth}{!}{%
\begin{tabular}{llcc}
\toprule
\textbf{Analysis Dimension} & \textbf{Configuration} & \textbf{NoveltyBench} & \textbf{IdeaBench} \\
\midrule
% PART I: Model Scale
\multicolumn{4}{l}{\textit{\textbf{I. Effect of Model Scale \& Paradigm}}} \\
Qwen3-1.7B & Vanilla & 4.82 & 0.39 \\
 & \cellcolor[rgb]{.867, .922, .969}+ PRISM & \cellcolor[rgb]{.867, .922, .969}\textbf{5.17} & \cellcolor[rgb]{.867, .922, .969}\textbf{0.65} \\
\cmidrule{1-4}
Qwen3-4B-Instruct & Vanilla & 5.05 & 0.63 \\
 & \cellcolor[rgb]{.867, .922, .969}+ PRISM & \cellcolor[rgb]{.867, .922, .969}\textbf{5.48} & \cellcolor[rgb]{.867, .922, .969}\textbf{0.93} \\
\cmidrule{1-4}
Qwen3-4B-Thinking & Vanilla & 3.14 & 0.70 \\
 & \cellcolor[rgb]{.867, .922, .969}+ PRISM & \cellcolor[rgb]{.867, .922, .969}\textbf{6.61} & \cellcolor[rgb]{.867, .922, .969}\textbf{0.85} \\

\midrule
\midrule

% PART II: Seed Count (Merged from Table 1)
\multicolumn{4}{l}{\textit{\textbf{II. Effect of Seed Count (gpt-4o-mini)}}} \\
Exploration Breadth & 3 Seeds & 3.61 & 0.57 \\
 & 8 Seeds & 2.45 & 0.52 \\
 & 15 Seeds & 3.67 & 0.41 \\

\midrule
\midrule

% PART III: Seed Source (Original Part II)
\multicolumn{4}{l}{\textit{\textbf{III. Effect of Lexical Seed Source}}} \\
Lexical Pool & General Pool (Standard) & 4.42 & 0.57 \\
 & Multi-Domain Pool & 4.41 & 0.61 \\
\bottomrule

\end{tabular}
\vspace*{-1pt}
}
\end{table}

\subsection{How Graph Helps}
\label{sec:analysis_combined}

The significant gains observed across creativity and discovery benchmarks raise a fundamental question: does the performance stem merely from accessing external information, or from the \emph{structured organization} of that information? We answer this by isolating the distinct contributions of topological structure, model capacity, and exploration breadth.

We first test the hypothesis that structure, not just retrieval volume, is the catalyst for creativity. To do this, we compare our full pipeline against the Search-Only baseline (Flat RAG), which utilizes the exact same retrieved chunks but lacks the graph construction phase. As shown in Table~\ref{tab:ablation_core}, the Full System significantly outperforms Flat RAG on both tasks (especially +16.33\% on IdeaBench). This confirms that raw information injection is insufficient; without structural organization, the model tends to treat dense external signals as stochastic noise and regress toward its prior mean. The graph structure is essential to contextualize these signals, enabling the model to internalize retrieved knowledge.

% \textbf{Scaling Effects and Cognitive Maturity.} 
% Beyond structure, we investigate how internal reasoning capacity interacts with the system. Using the \texttt{Qwen3} family (Table~\ref{tab:model_scale}), we observe that while gains are universal, reasoning-optimized models (\texttt{4B-Thinking}) show a non-linear leap (+110.5\% in Novelty). This suggests a synergistic relationship: the epistemic graph provides a complex reasoning substrate that ``Thinking'' models are uniquely equipped to navigate and exploit.

% \textbf{The Entropy-Coherence Trade-off.}
% Finally, we examine the stability of this exploration. As shown in Table \ref{tab:sensitivity_combined}, increasing seeds from 3 to 8 causes a temporary performance dip due to semantic noise. However, performance recovers at 15 seeds as the graph becomes dense enough to re-establish global coherence. This underscores that PRISM acts as a stabilizing constraint, allowing for high-entropy exploration without losing the logical thread.\dy{@Guancheng}

\subsection{Impact of Model Scale and Training Paradigm}
To decouple parameter count from training methodology, we evaluate the \texttt{Qwen3} family by comparing three variants: the lightweight \texttt{1.7B}, the standard \texttt{4B-Instruct}, and the reasoning-enhanced \texttt{4B-Thinking}.

Table~\ref{tab:sensitivity_combined} reports the performance breakdown. While gains are universal, the magnitude varies significantly. The non-linear jump in \texttt{Qwen3-4B-Thinking} (doubling the NoveltyBench score) reveals a synergistic leap: the epistemic graph provides the \textit{lived experience} that reasoning models possess the inherent capability to better leverage. Conversely, the improvements on the \texttt{1.7B} model confirm that the graph effectively functions as an external cognitive prosthesis even for smaller models with limited intrinsic capacity.

\subsection{Effect of Multi-Seed Exploration Breadth}
In Section~\ref{sec:rarebench}, we emphasized the importance of seed \emph{quality} (intent). Here, we examine the impact of seed \emph{quantity}. The number of initial seeds controls the breadth of the entry points into the epistemic graph. We compare configurations with 3, 8, and 15 seeds using \texttt{gpt-4o-mini}.

As shown in Table ~\ref{tab:sensitivity_combined}, the relationship is non-monotonic. Increasing from 3 to 15 seeds recovers performance, but the intermediate 8-seed configuration shows a significant drop (-32.14\% on NoveltyBench). This counter-intuitive finding suggests that moderate expansion may introduce semantic noise, as the graph structure is not yet dense enough to organize effectively. A larger seed set (15) appears to provide sufficient connectivity to re-establish coherence. This highlights a trade-off: distinct performance profiles emerge at different levels of ``cognitive load,'' suggesting that optimal seed counts may be task-dependent.

\subsection{Influence of Domain-Specific Lexical Pools}
Finally, we investigate the semantic source of the seeds. We compare the general-purpose noun pool used in our main experiments against an expanded pool incorporating domain-specific vocabularies (e.g., biology, physics).

Table \ref{tab:sensitivity_combined} reveals a nuanced impact: domain-specific seeds provide a clear benefit for scientific discovery (IdeaBench, +5.2\%) but show negligible impact on general creativity (NoveltyBench). This indicates that unconstrained semantic expansion is not universally beneficial; rather, the diversity of the input space must be aligned with the specific reasoning context of the downstream task.

\section{Related Work}
\label{sec:related}
\noindent \textbf{Diversity and The Artificial Hivemind}
The phenomenon of mode collapse or Artificial Hivemind has emerged as a critical concern, where LLMs trained on recursive synthetic data converge towards low-entropy, homogenized distributions~\cite{jiang2025artificial}. To mitigate this, existing research has primarily explored two avenues. The first focuses on inference-time stochasticity, employing sampling strategies like high temperature~\cite{ficler2017controlling}, nucleus sampling~\cite{holtzman2019curious}, etc. 
The second involves training-time interventions, such as diversity-aware fine-tuning or objective functions that penalize mode collapse~\cite{franceschelli2025creativity,ismayilzada2025creative}. While these methods successfully increase statistical variance, they often conflate randomness with creativity~\cite{jiang2025artificial}. PRISM takes a different approach: rather than manipulating probability logits or model weights (a strategy we deliberately avoid due to the optimization paradox and safety risks detailed in Appendix~\ref{appendix:dilemma_of_training}), we address the root cause: the homogenization of the epistemic context. By structurally differentiating the information intake via Wild Search, we induce semantic divergence that is grounded in evidence rather than stochastic noise.

\noindent \textbf{Retrieval-Augmented Generation}
RAG and GraphRAG serve as a promising architecture for extending parametric memory~\cite{gao2023retrieval,edge2024local}. PRISM builds upon these techniques, yet repurposes them to operationalize the \textit{Nurture} component of our framework. While standard RAG typically targets factual retrieval to answer user queries~\cite{singal2024evidence,yang2024rag}, we treat retrieval as the vehicle for Experiencing—gathering heterogeneous information to simulate a unique research trajectory. Consequently, the graph component in PRISM functions not as a static knowledge base for fact-checking, but as the mechanism for Cognitive Internalization. By constructing an on-the-fly epistemic graph, our system transforms raw retrieved experiences into a dynamic reasoning substrate. This shift moves beyond optimizing for immediate relevance, using the RAG pipeline instead to prioritize semantic dispersion and concept bridging, thereby facilitating the emergence of a distinct, individualized perspective.

\noindent \textbf{Agentic Memory and Cognitive Simulation} 
Our work also aligns with the growing field of agentic cognitive architectures. Approaches like Persona Prompting~\cite{argyle2023out} or Role-Play~\cite{kong2024better} utilize extensive system prompts to simulate specific perspectives. While effective for stylistic adaptation, these methods often result in superficial simulations lacking deep, domain-specific grounding. More advanced architectures, such as Generative Agents~\cite{park2023generative} and MemGPT~\cite{packer2023memgpt}, introduce persistent memory streams to facilitate long-term behavioral consistency. PRISM extends this lineage by focusing specifically on epistemic individuation. Unlike memory streams that simulate social behavior or biographical history, PRISM’s graph models the cognitive trajectory of a research journey. It functions as an Extended Mind~\cite{clark1998extended}, where the external graph structure actively shapes the internal reasoning process, enabling the emergence of distinct perspectives without relying on rigid role definitions.

% A common approach to eliciting diversity is Persona Prompting~\cite{argyle2023out} or Role-Play (e.g., "Act as a biologist"). While this can shift the output distribution, it often suffers from caricature simulation: the model mimics the surface-level style of a persona without possessing the corresponding unique knowledge base, yielding simulated diversity without genuine reasoning divergence. PRISM replaces this superficial prompting with structural nurture. Instead of telling the model who to be, we provide it with the unique experiences (via the Epistemic Graph~\cite{nikooroo2025belief}) that naturally lead to a distinct perspective. This ensures that the resulting diversity is a genuine product of different reasoning paths, not just a stylistic filter.

% ==================================================
\section{Future Work}
We outline future directions across two dimensions, expecting the community to extend this work based on available resources and expertise. 
For institutions with substantial computational capacity, the architectural challenge lies in (1) evolving Epistemic Graphs, transitioning the system from episodic retrieval to continuous, lifelong learning; and (2) expanding exploration into large-scale, active curiosity-driven strategies.
Conversely, for interdisciplinary experts, PRISM offers a ready-to-use framework for science discovery. We expect researchers to deploy PRISM across diverse scientific verticals, validating its utility or even benefit AI for Science research.

% A key direction is to reduce inference latency by enabling parallel execution and incremental graph updates, thereby lowering end-to-end runtime. We also plan to develop query-aware exploration control that adaptively allocates the number of seeds, retrieval budget, and graph size based on query complexity. To improve robustness under noisy evidence, we will incorporate source reliability modeling and conflict-aware aggregation to mitigate error propagation from imperfect retrieval. Finally, we will strengthen creativity evaluation by complementing automatic metrics with expert human studies and task-based protocols that better reflect real-world usefulness.

% ==================================================
\section{Conclusion}
\label{sec:conclusion}
In this work, we proposed breaking the Artificial Hivemind by shifting monolithic, static inference via Epistemic Evolution -- an ecosystem where AI share a common foundation but diverge through unique experiences. PRISM validates this paradigm, achieving state-of-the-art performance across creativity and diversity benchmarks while maintaining semantic meaningfulness. Looking forward, as the technique for digital experiencing and cognitive internalization keep improving alongside model scaling, we expect that the Epistemic Evolution paradigm for \textbf{Pluralistic AI} will serve as the differentiating factor, transforming generic assistants into distinct, capable engines of AI-augmented scientific discovery.

\section*{Impact Statement}

This paper presents work whose goal is to advance the field of machine learning by moving beyond monolithic model consensus toward Pluralistic AI. The widespread convergence of Large Language Models (LLMs) towards an Artificial Hivemind poses significant societal risks, including the erosion of cultural diversity, the stagnation of scientific discovery, and the marginalization of long-tail knowledge. By introducing PRISM to enable distinct ``epistemic trajectories," our work aims to restore the diversity of thought necessary for creative exploration and complex problem-solving.

Societal Benefits: The immediate positive impact of this work is demonstrated in high-stakes domains such as healthcare. As shown in our evaluation on RareBench, breaking the ``conservative bias" of standard models can significantly improve the diagnosis of rare diseases that are often overlooked by consensus-seeking algorithms. Broadly, enabling AI to explore diverse, grounded perspectives can accelerate scientific discovery by surfacing novel hypotheses that statistically average models might discard.

Ethical Considerations and Safety: We acknowledge that mechanisms designed to increase response diversity carry inherent risks. \begin{itemize} \item \textbf{Safety Guardrails:} Techniques that alter inference-time context to induce ``unique personalities" could potentially weaken the safety alignment (RLHF) of the base model. However, because PRISM operates strictly at inference time without modifying model weights, it retains the underlying safety training of the base model. We emphasize that the epistemic graph acts as a grounding constraint, distinct from unconstrained jailbreaking. \item \textbf{Truthfulness vs. Hallucination:} Inducing novelty runs the risk of inducing hallucination. Our approach mitigates this by requiring that divergence be grounded in retrieved evidence (the \textit{Nurture} component) rather than pure stochastic sampling. Nevertheless, the quality of the Wild Search is critical; if the system retrieves misinformation, the graph may internalize it. Future deployment must ensure robust filtering of the retrieval corpus. \end{itemize}

Ultimately, this work advocates for a shift in the human-AI relationship: from treating the AI as a singular oracle of truth to viewing it as a diverse ecosystem of collaborators. We believe this shift encourages critical thinking and active verification, essential skills in the era of generative AI.

\bibliography{example_paper}

@article{jiang2025artificial,
  title={Artificial hivemind: The open-ended homogeneity of language models (and beyond)},
  author={Jiang, Liwei and Chai, Yuanjun and Li, Margaret and Liu, Mickel and Fok, Raymond and Dziri, Nouha and Tsvetkov, Yulia and Sap, Maarten and Albalak, Alon and Choi, Yejin},
  journal={arXiv preprint arXiv:2510.22954},
  year={2025}
}

@article{argyle2023out,
  title={Out of one, many: Using language models to simulate human samples},
  author={Argyle, Lisa P and Busby, Ethan C and Fulda, Nancy and Gubler, Joshua R and Rytting, Christopher and Wingate, David},
  journal={Political Analysis},
  volume={31},
  number={3},
  pages={337--351},
  year={2023},
  publisher={Cambridge University Press}
}

@inproceedings{park2023generative,
  title={Generative agents: Interactive simulacra of human behavior},
  author={Park, Joon Sung and O'Brien, Joseph and Cai, Carrie Jun and Morris, Meredith Ringel and Liang, Percy and Bernstein, Michael S},
  booktitle={Proceedings of the 36th annual acm symposium on user interface software and technology},
  pages={1--22},
  year={2023}
}

@article{nikooroo2025belief,
  title={Belief Graphs with Reasoning Zones: Structure, Dynamics, and Epistemic Activation},
  author={Nikooroo, Saleh and Engel, Thomas},
  journal={arXiv preprint arXiv:2510.10042},
  year={2025}
}

@inproceedings{guo2025ideabench,
  title={Ideabench: Benchmarking large language models for research idea generation},
  author={Guo, Sikun and Shariatmadari, Amir Hassan and Xiong, Guangzhi and Huang, Albert and Kim, Myles and Williams, Corey M and Bekiranov, Stefan and Zhang, Aidong},
  booktitle={Proceedings of the 31st ACM SIGKDD Conference on Knowledge Discovery and Data Mining V. 2},
  pages={5888--5899},
  year={2025}
}

@article{zhang2025noveltybench,
  title={NoveltyBench: Evaluating Language Models for Humanlike Diversity},
  author={Zhang, Yiming and Diddee, Harshita and Holm, Susan and Liu, Hanchen and Liu, Xinyue and Samuel, Vinay and Wang, Barry and Ippolito, Daphne},
  journal={arXiv preprint arXiv:2504.05228},
  year={2025}
}

@inproceedings{chen2024rarebench,
  title={RareBench: can LLMs serve as rare diseases specialists?},
  author={Chen, Xuanzhong and Mao, Xiaohao and Guo, Qihan and Wang, Lun and Zhang, Shuyang and Chen, Ting},
  booktitle={Proceedings of the 30th ACM SIGKDD conference on knowledge discovery and data mining},
  pages={4850--4861},
  year={2024}
}

@article{luchins1942mechanization,
  title={Mechanization in problem solving: The effect of Einstellung.},
  author={Luchins, Abraham S},
  journal={Psychological monographs},
  volume={54},
  number={6},
  pages={i},
  year={1942},
  publisher={American Psychological Association}
}

@book{bartlett1995remembering,
  title={Remembering: A study in experimental and social psychology},
  author={Bartlett, Frederic Charles},
  year={1995},
  publisher={Cambridge university press}
}

@article{clark1998extended,
  title={The extended mind},
  author={Clark, Andy and Chalmers, David},
  journal={analysis},
  volume={58},
  number={1},
  pages={7--19},
  year={1998},
  publisher={JSTOR}
}

@article{hong2004groups,
  title={Groups of diverse problem solvers can outperform groups of high-ability problem solvers},
  author={Hong, Lu and Page, Scott E},
  journal={Proceedings of the National Academy of Sciences},
  volume={101},
  number={46},
  pages={16385--16389},
  year={2004},
  publisher={National Academy of Sciences}
}

@article{wang2024mmlu,
  title={Mmlu-pro: A more robust and challenging multi-task language understanding benchmark},
  author={Wang, Yubo and Ma, Xueguang and Zhang, Ge and Ni, Yuansheng and Chandra, Abhranil and Guo, Shiguang and Ren, Weiming and Arulraj, Aaran and He, Xuan and Jiang, Ziyan and others},
  journal={Advances in Neural Information Processing Systems},
  volume={37},
  pages={95266--95290},
  year={2024}
}

@article{huang2025winning,
  title={Winning gold at imo 2025 with a model-agnostic verification-and-refinement pipeline},
  author={Huang, Yichen and Yang, Lin F},
  journal={arXiv preprint arXiv:2507.15855},
  year={2025}
}

@article{kirk2023understanding,
  title={Understanding the effects of rlhf on llm generalisation and diversity},
  author={Kirk, Robert and Mediratta, Ishita and Nalmpantis, Christoforos and Luketina, Jelena and Hambro, Eric and Grefenstette, Edward and Raileanu, Roberta},
  journal={arXiv preprint arXiv:2310.06452},
  year={2023}
}

@article{comanici2025gemini,
  title={Gemini 2.5: Pushing the frontier with advanced reasoning, multimodality, long context, and next generation agentic capabilities},
  author={Comanici, Gheorghe and Bieber, Eric and Schaekermann, Mike and Pasupat, Ice and Sachdeva, Noveen and Dhillon, Inderjit and Blistein, Marcel and Ram, Ori and Zhang, Dan and Rosen, Evan and others},
  journal={arXiv preprint arXiv:2507.06261},
  year={2025}
}

@article{singh2025openai,
  title={Openai gpt-5 system card},
  author={Singh, Aaditya and Fry, Adam and Perelman, Adam and Tart, Adam and Ganesh, Adi and El-Kishky, Ahmed and McLaughlin, Aidan and Low, Aiden and Ostrow, AJ and Ananthram, Akhila and others},
  journal={arXiv preprint arXiv:2601.03267},
  year={2025}
}

@article{ouyang2022training,
  title={Training language models to follow instructions with human feedback},
  author={Ouyang, Long and Wu, Jeffrey and Jiang, Xu and Almeida, Diogo and Wainwright, Carroll and Mishkin, Pamela and Zhang, Chong and Agarwal, Sandhini and Slama, Katarina and Ray, Alex and others},
  journal={Advances in neural information processing systems},
  volume={35},
  pages={27730--27744},
  year={2022}
}

@article{yue2025does,
  title={Does reinforcement learning really incentivize reasoning capacity in llms beyond the base model?},
  author={Yue, Yang and Chen, Zhiqi and Lu, Rui and Zhao, Andrew and Wang, Zhaokai and Song, Shiji and Huang, Gao},
  journal={arXiv preprint arXiv:2504.13837},
  year={2025}
}

@article{yang2025many,
  title={Many Minds from One Model: Bayesian Transformers for Population Intelligence},
  author={Yang, Diji and Zhang, Yi},
  journal={arXiv preprint arXiv:2512.25063},
  year={2025}
}

@article{franceschelli2025creativity,
  title={On the creativity of large language models},
  author={Franceschelli, Giorgio and Musolesi, Mirco},
  journal={AI \& society},
  volume={40},
  number={5},
  pages={3785--3795},
  year={2025},
  publisher={Springer}
}

@article{ficler2017controlling,
  title={Controlling linguistic style aspects in neural language generation},
  author={Ficler, Jessica and Goldberg, Yoav},
  journal={arXiv preprint arXiv:1707.02633},
  year={2017}
}

@article{holtzman2019curious,
  title={The curious case of neural text degeneration},
  author={Holtzman, Ari and Buys, Jan and Du, Li and Forbes, Maxwell and Choi, Yejin},
  journal={arXiv preprint arXiv:1904.09751},
  year={2019}
}

@article{ismayilzada2025creative,
  title={Creative preference optimization},
  author={Ismayilzada, Mete and Laverghetta, Antonio and Luchini, Simone A and Patel, RN and Bosselut, Antoine and Van Der Plas, Lonneke and Beaty, Roger E},
  journal={Findings of the Association for Computational Linguistics: EMNLP 2025},
  pages={9580--9609},
  year={2025},
  publisher={Association for Computational Linguistics}
}

@article{gao2023retrieval,
  title={Retrieval-augmented generation for large language models: A survey},
  author={Gao, Yunfan and Xiong, Yun and Gao, Xinyu and Jia, Kangxiang and Pan, Jinliu and Bi, Yuxi and Dai, Yixin and Sun, Jiawei and Wang, Haofen and Wang, Haofen},
  journal={arXiv preprint arXiv:2312.10997},
  volume={2},
  number={1},
  year={2023}
}

@article{edge2024local,
  title={From local to global: A graph rag approach to query-focused summarization},
  author={Edge, Darren and Trinh, Ha and Cheng, Newman and Bradley, Joshua and Chao, Alex and Mody, Apurva and Truitt, Steven and Metropolitansky, Dasha and Ness, Robert Osazuwa and Larson, Jonathan},
  journal={arXiv preprint arXiv:2404.16130},
  year={2024}
}

@article{zeng2025worse,
  title={Worse than zero-shot? a fact-checking dataset for evaluating the robustness of rag against misleading retrievals},
  author={Zeng, Linda and Gupta, Rithwik and Motwani, Divij and Zhang, Yi and Yang, Diji},
  journal={arXiv preprint arXiv:2502.16101},
  year={2025}
}

@inproceedings{yang2024rag,
  title={Im-rag: Multi-round retrieval-augmented generation through learning inner monologues},
  author={Yang, Diji and Rao, Jinmeng and Chen, Kezhen and Guo, Xiaoyuan and Zhang, Yawen and Yang, Jie and Zhang, Yi},
  booktitle={Proceedings of the 47th International ACM SIGIR Conference on Research and Development in Information Retrieval},
  pages={730--740},
  year={2024}
}

@inproceedings{kong2024better,
  title={Better zero-shot reasoning with role-play prompting},
  author={Kong, Aobo and Zhao, Shiwan and Chen, Hao and Li, Qicheng and Qin, Yong and Sun, Ruiqi and Zhou, Xin and Wang, Enzhi and Dong, Xiaohang},
  booktitle={Proceedings of the 2024 Conference of the North American Chapter of the Association for Computational Linguistics: Human Language Technologies (Volume 1: Long Papers)},
  pages={4099--4113},
  year={2024}
}

@article{packer2023memgpt,
  title={MemGPT: Towards LLMs as Operating Systems.},
  author={Packer, Charles and Fang, Vivian and Patil, Shishir\_G and Lin, Kevin and Wooders, Sarah and Gonzalez, Joseph\_E},
  year={2023},
  publisher={ArXiv}
}

@article{zhang2019bertscore,
  title={Bertscore: Evaluating text generation with bert},
  author={Zhang, Tianyi and Kishore, Varsha and Wu, Felix and Weinberger, Kilian Q and Artzi, Yoav},
  journal={arXiv preprint arXiv:1904.09675},
  year={2019}
}

@inproceedings{singal2024evidence,
  title={Evidence-backed fact checking using RAG and few-shot in-context learning with LLMs},
  author={Singal, Ronit and Patwa, Pransh and Patwa, Parth and Chadha, Aman and Das, Amitava},
  booktitle={Proceedings of the Seventh Fact Extraction and VERification Workshop (FEVER)},
  pages={91--98},
  year={2024}
}
\bibliographystyle{icml2026}

\newpage
\appendix
\onecolumn

\section{Discussion: The Cognitive Dynamics of Pluralism}
\label{appendix:pluralish}
Recent work identifies the Artificial Hivemind as a dual threat: intra-model repetition, where models lock into repetitive loops, and inter-model homogeneity, where diverse models converge on identical outputs~\citep{jiang2025artificial}. We argue that PRISM addresses these failures not merely as engineering constraints, but by modeling the cognitive mechanisms of individuation.

\paragraph{Breaking Intra-model via Epistemic Schemas}
Intra-model repetition mirrors the human phenomenon of cognitive fixation or mental sets (Einstellung effect), where prior experience blinds a solver to novel solutions~\citep{luchins1942mechanization}. Standard LLMs, constrained by the ``average'' of their training data, exhibit a static, global fixation. PRISM breaks this by injecting distinct epistemic seeds—effectively initializing a unique schema~\citep{bartlett1995remembering} for each inference pass. Consider a researcher deeply immersed in Reinforcement Learning: they interpret daily challenges, from traffic to diet, through the lens of reward maximization. Similarly, a PRISM graph seeded with domain-specific concepts forces the model to adopt a temporary, specialized ``personality," exploring the solution space through a distinctive conceptual lens rather than a generic means.

\paragraph{The Extended Mind and Dynamic Trajectories}
Human cognition is not static; it evolves through accumulation. The insights of a young Isaac Newton differ profoundly from those of the elder master of the Mint, separated by decades of experience. Static context windows cannot capture this evolution, but our dynamic epistemic graph can. Drawing on the Extended Mind Thesis~\citep{clark1998extended}, which posits that external structures (like notebooks or graphs) function as constitutive parts of the cognitive process, PRISM's graph serves as an evolving external memory. As the wild search progresses, the graph assimilates new nodes and accommodates its structure, simulating how \textit{nurture} (experience) dynamically reshapes the model's reasoning trajectory in real-time.

\paragraph{From inter-model homogeneity to Collective Intelligence}
Finally, the Hivemind's most critical failure is the illusion that a perfect alignment yields the best collective outcome. Complex systems theory suggests the opposite: the Diversity Prediction Theorem demonstrates that a crowd of diverse problem solvers often outperforms a crowd of high-ability but homogeneous experts~\citep{hong2004groups}. By equipping each model instance with a unique, graph-mediated Nurture, PRISM transforms a monolithic array of clones into a pluralistic society of agents. This suggests that the future of machine intelligence lies not in training a single, perfect sage, but in orchestrating a diverse ecosystem of distinct cognitive individuals.

\paragraph{Motivation: From Nature to Nurture.}
The design of the Cognitive Explosion module is rooted in our vision of the \textbf{Life-long Epistemic Graph}---a dynamic, evolving storage system that records an agent's entire trajectory, including tool invocations, human-AI interactions, and external knowledge acquisition. We posit that while pre-training provides the model's ``Nature,'' true intelligence requires \textit{Nurture} through individualized experiences.

However, capturing a full lifecycle of interaction is computationally prohibitive for a single study. As a proof-of-concept, we employ \textit{Wild Search} to simulate these formative experiences. The stochastic sampling of lexical units mimics the serendipitous nature of human developmental milestones---much like the random yet decisive influence of choosing a university major or a first professional internship. By injecting these stochastic perturbations, we ensure that the Epistemic Graph begins with a unique initialization, effectively individualizing the exploration paths and knowledge backgrounds of different model instances.

\section{The Dilemma of Training for Diversity}
\label{appendix:dilemma_of_training}
While re-training or fine-tuning models to encourage diversity seems intuitive, it faces three fundamental hurdles. First, the Optimization Paradox: Creativity is inherently subjective and sparse. Unlike accuracy or safety, defining a robust loss function for interesting divergence without devolving into hallucination is notoriously difficult. Second, the Exploration-Exploitation Trade-off: Current alignment paradigms (SFT/RLHF/RLVR) are optimized for exploitation—reliably generate the correct or safest answer. Forcing the model to explore the long-tail distribution during training often degrades its general instruction-following capabilities. Third, the Cost and Safety Risk: Retraining foundation models requires prohibitive computational resources and necessitates re-verifying the entire safety pipeline, as distinct personalities may inadvertently bypass existing well-tested safety guardrails. For above reasons, in this work, we deliberately avoided any tuning to the model's weights, instead choosing to alter the model's behavior during the inference phase.

\section{Appendix: Experimental Configuration}
\label{sec:appendix_config}

This appendix provides a comprehensive description of the PRISM system's internal configuration and the evaluation protocols used across our experiments. Our goal is to ensure full reproducibility and to clarify the design choices made during the development of the pipeline.

\subsection{PRISM System Implementation}
\label{sec:prism_config_refined}

The PRISM framework operationalizes creative inference through a structured multi-stage graph construction process.

\paragraph{Search and Data Sampling.} 
For the retrieval phase, we utilize ClueWeb22 to perform large-scale ``Wild Search" for each input query. Given the volume of retrieved documents, we segment the text into chunks of approximately 400 tokens. To control computational overhead while introducing necessary stochastic diversity into the epistemic space, we randomly sample a subset of these chunks. Following preliminary experiments, the chunk sampling size is fixed at 8 for all experiments.

\paragraph{Graph Construction and Node Sampling.} 
The system constructs an epistemic graph by first initializing each query with 3 lexical seeds. The resulting graph consists of two primary node types: \textbf{Context Nodes}, extracted from the user query to represent structural constraints and abstract ideas, and \textbf{Spark Nodes}, extracted from the retrieved chunks to act as creative triggers. To ensure efficient edge generation during the bridging phase, we limit the number of spark nodes to 7 via random sampling. Creative edges are then generated by applying bridging operators to pairwise combinations of context and sampled spark nodes.

\paragraph{Inference Temperature Schedule.} 
We employ a stage-specific temperature schedule to balance deterministic precision with exploratory creativity. For \textbf{Context Extraction}, we use $T=0.0$ to ensure that user constraints are modeled with high fidelity. \textbf{Spark Extraction} adopts a moderate $T=0.3$ to allow for diversity in inspiration units. Finally, the \textbf{Bridging and Creative Edge Generation} phase employs a high temperature of $T=1.2$, facilitating the discovery of novel associations within the constrained epistemic space.

\subsection{NovelBench Evaluation Protocol}
\label{sec:novelbench_config_refined}

We evaluate PRISM using the NovelBench benchmark, strictly adhering to its original protocol to maintain fair comparisons.

\paragraph{Generation and Scoring Settings.} 
All base models and PRISM-enhanced variants generate responses using a decoding temperature of 1.0. For each prompt, we sample 10 independent responses to assess distributional diversity. We set the user patience parameter to $p=0.8$. Utility scores are measured using the \texttt{Skywork-Reward-Gemma-2-27B-v0.2} reward model, while \textit{Distinct-k} scores are calculated via the official \texttt{deberta-v3-large} equivalence detection model.

\paragraph{Extended Results and Discussion.} 
Table~\ref{tab:complete_bench_results} provides the complete evaluation results. The upper section contains reference performance data for various model families as reported in the original NovelBench paper, while the lower section illustrates the gains achieved by our PRISM system.

\begin{table}[htbp]
\centering
\small
\caption{Comprehensive NovelBench results. The upper section lists updated reference baselines for major model families. The lower section shows our testbed results, where $\hookrightarrow$ denotes the PRISM-enhanced version. All models are evaluated on \textit{Distinct} (Novelty) and \textit{Utility}.}
\label{tab:complete_bench_results}
\begin{tabular}{l cc}
\toprule
\textbf{Model} & \textbf{Distinct} & \textbf{Utility} \\
\midrule
\textit{Baselines (Reference Data)} & & \\
Claude-3.5 Haiku & 1.94 & 2.50 \\
Claude-3.5 Sonnet & 1.76 & 2.36 \\
Claude-3 Opus & 2.04 & 2.67 \\
\addlinespace
gpt-4o & 2.88 & 3.27 \\
\addlinespace
gemini-1.5-pro & 1.85 & 2.73 \\
gemini-2.0-flash-lite & 2.83 & 3.20 \\
gemini-2.0-flash & 2.81 & 3.17 \\
gemini-2.0-pro & 2.25 & 2.64 \\
\addlinespace
command-r7b & 3.58 & 3.35 \\
command-r & 2.68 & 2.98 \\
command-r-plus & 2.79 & 3.08 \\
\addlinespace
gemma-2-2b-it & 5.66 & 4.63 \\
gemma-2-9b-it & 3.25 & 3.93 \\
gemma-2-27b-it & 3.03 & 3.77 \\
\addlinespace
Llama-3.2-1B & 6.74 & 2.81 \\
Llama-3.2-3B & 5.10 & 3.24 \\
Llama-3.1-8B & 5.24 & 3.76 \\
Llama-3.3-70B & 2.49 & 2.87 \\
Llama-3.1-405B & 3.20 & 3.39 \\

\midrule
\textit{Main Results (Our Experiments)} & & \\
Qwen3-4B-Instruct & 3.09 & 2.24 \\
~~~~$\hookrightarrow$ \textbf{PRISM} & \textbf{4.48} & 2.00 \\
\addlinespace
CrPO-sft-LLaMA-3.1 & 7.35 & 3.38 \\
~~~~$\hookrightarrow$ \textbf{PRISM} & \textbf{7.67} & 2.95 \\
\addlinespace
gpt-4o-mini & 2.65 & 3.11 \\
~~~~$\hookrightarrow$ \textbf{PRISM} & \textbf{3.41} & 2.08 \\
\bottomrule
\end{tabular}
\end{table}

While we observe a moderate decrease in utility scores, this behavior is expected. Reward models are typically trained to align with average human preferences, which inherently favors consensus-based, ``mean-regressive" outputs. Such objectives are often misaligned with high novelty, as creative outputs may depart from common patterns. However, the moderate nature of the utility degradation suggests that PRISM successfully maintains semantic coherence through its epistemic graph constraints, achieving a robust balance between exploration and relevance.

\subsection{IdeaBench Evaluation Protocol}
\label{sec:ideabench_config}

The evaluation on IdeaBench follows a structured pipeline designed to assess the quality of scientific hypothesis generation across multiple dimensions. 

\paragraph{Evaluation Pipeline.} 
For each target paper, the system first collects relevant reference abstracts as background context. Using a unified prompt template, the language model under test generates $n=3$ research hypotheses. Simultaneously, GPT-4o is employed to rewrite the original target paper's abstract into a standardized hypothesis format, serving as the human-level baseline. This human hypothesis is then combined with the $n$ model-generated hypotheses to form a candidate pool for subsequent evaluation.

\paragraph{Semantic Consistency and Idea Overlap.} 
To measure the alignment between generated ideas and the ground truth, two primary metrics are utilized:
\begin{itemize}
    \item \textbf{BERTScore (F1):} This metric computes the semantic similarity between each generated hypothesis and the target abstract using contextual embeddings from a pre-trained BERT model. For each paper, the maximum BERTScore among the $n$ generated candidates is selected. The 80th percentile of these maximum scores across the entire test set is reported as the final semantic consistency score.
    \item \textbf{Idea Overlap:} GPT-4o acts as an automatic scorer to evaluate the content overlap between each generated hypothesis and the target abstract. Ratings are assigned on a scale of 1 to 10, accompanied by brief justifications. The highest overlap score among the $n$ candidates for each paper is recorded.
\end{itemize}

\paragraph{Ranking-based Quality Assessment.} 
A blind ranking paradigm is adopted to evaluate core quality. The consolidated candidate set (one human baseline and $n$ model-generated hypotheses) is presented to GPT-4o. The scorer ranks all candidates based on a specific quality dimension $q$ (e.g., \textit{Novelty} or \textit{Feasibility}) without knowledge of their sources. Let $r_{i|q} \in \{1, \dots, n+1\}$ denote the rank of the human baseline for the $i$-th paper under dimension $q$. The \textbf{Insight Score} is defined as:
\begin{equation}
I(\text{LLM}, q) = \frac{1}{m} \sum_{i=1}^{m} \frac{r_{i|q} - 1}{n},
\end{equation}
where $m$ represents the total number of papers. This score characterizes the overall relative advantage of the model-generated ideas compared to the human baseline in dimension $q$.

\paragraph{Experimental Setup.} 
 For the PRISM system and all baseline models, we strictly follow all hyperparameters from the evaluation codebase of IdeaBench~\cite{guo2025ideabench} to ensure the fair comparison. Results are aggregated and averaged across the entire test set, reporting the BERTScore~\cite{zhang2019bertscore}, Idea Overlap, and Insight Scores for both Novelty and Feasibility.

\begin{table}[ht]
\centering
\caption{Comprehensive results on IdeaBench. Our system (PRISM) significantly boosts the Novelty Insight Score compared to various baselines and model scenarios. All scores are rounded to two decimal places.}
\label{tab:combined_results}
\small 
\setlength{\tabcolsep}{12pt} 
\begin{tabular}{lccc}
\toprule
\textbf{Model / Scenario} & \textbf{SemSim $\downarrow$} & \textbf{Overlap $\downarrow$} & \textbf{Novelty $\uparrow$} \\
\midrule
\textit{Main Results (Our Experiments)} & & & \\
Qwen3-4B-Instruct & 0.60 & 6 & 0.72 \\
\quad $\hookrightarrow$ + PRISM & 0.56 & 5 & \textbf{0.96} \\
\addlinespace
gpt-4o-mini & 0.61 & 7 & 0.45 \\
\quad $\hookrightarrow$ + PRISM & 0.59 & 7 & \textbf{0.65} \\
\midrule
\textit{Baselines (Reference Data)} & & & \\
Llama 3.1 70B-Instruct (low) & 0.59 & 7 & 0.62 \\
Llama 3.1 70B-Instruct (high) & 0.60 & 8 & 0.60 \\
\addlinespace
Llama 3.1 405B-Instruct (low) & 0.57 & 8 & 0.65 \\
Llama 3.1 405B-Instruct (high) & 0.59 & 8 & 0.68 \\
\addlinespace
Gemini 1.5 Flash (low) & 0.59 & 7 & 0.43 \\
Gemini 1.5 Flash (high) & 0.59 & 8 & 0.57 \\
\addlinespace
Gemini 1.5 Pro (low) & 0.59 & 6 & 0.51 \\
Gemini 1.5 Pro (high) & 0.60 & 7 & 0.65 \\
\addlinespace
GPT-3.5 Turbo (low) & 0.61 & 8 & 0.40 \\
GPT-3.5 Turbo (high) & 0.62 & 8 & 0.20 \\
\addlinespace
GPT-4o Mini (low) & 0.61 & 7 & 0.45 \\
GPT-4o Mini (high) & 0.62 & 8 & 0.53 \\
\addlinespace
GPT-4o (low) & 0.60 & 7 & 0.61 \\
GPT-4o (high) & 0.61 & 8 & 0.77 \\
\bottomrule
\end{tabular}
\end{table}

\paragraph{Interpretation of Semantic Similarity and Overlap Scores.}
We note that the presentation of semantic similarity (SemSim) and idea overlap scores in Table~\ref{tab:combined_results} differs slightly from the original reporting convention in IdeaBench.
Specifically, in our setting, lower SemSim and Overlap values indicate stronger performance.

This design choice reflects the primary objective of our task, which is to encourage the generation of genuinely novel and independent research ideas.
Since both BERTScore-based semantic similarity and LLM-evaluated overlap measure the closeness between generated hypotheses and the target paper, lower scores correspond to ideas that are more distant from existing work, and thus potentially more innovative.

Importantly, lower similarity and overlap in our results do not indicate arbitrary or irrelevant generations.
As evidenced by the consistently high Novelty and Feasibility Insight Scores, our system maintains semantic coherence and practical plausibility while exploring more distant regions of the idea space.
This suggests that PRISM is able to balance creativity and feasibility, producing ideas that are both novel and well-grounded rather than speculative or disconnected.

Overall, this reversed interpretation of SemSim and Overlap aligns with our goal of evaluating controlled divergence from existing literature, highlighting the capability of our system to generate innovative yet actionable research hypotheses.

\subsection{Artificial Hivemind Configuration}
\label{sec:hivemind_config}

The \textit{Artificial Hivemind} experiment is conducted to analyze the collective behavior and distributional consistency of the models. The setup focuses on generating a high-density sample space to ensure statistical reliability.
\paragraph{Prompt Selection and Sampling.} A curated set of 15 representative prompts is utilized for this evaluation. To accurately capture the output distribution of each candidate model, 50 independent responses are sampled for every prompt. This results in a total of 750 generated samples per model ($15 \text{ prompts} \times 50 \text{ samples}$), providing a robust basis for analyzing response diversity and consensus.
\paragraph{Model Hyperparameters.} Consistency across models is maintained by setting a uniform decoding temperature of $T=0.7$. For all tasks requiring semantic vectorization—including the calculation of centroid embeddings or intra-group similarity—the \texttt{text-embedding-3-small} model is employed as the fixed embedding backbone.

% ------------------------------------------------

% \subsection{Additional Materials}

% The theoretical formulation of the ``Nature vs.\ Nurture'' paradigm is provided in Section~\ref{sec:theory} for completeness.

\subsection{Rarebench Experimental Setup}
\label{sec:rareexp_setup}

\paragraph{Dataset.}
We evaluate our framework on the RAMEDIS subset of RareBench~\cite{chen2024rarebench}, a challenging benchmark for rare disease diagnosis. The dataset comprises 624 patient cases, where phenotype descriptions are mapped to standard Human Phenotype Ontology (HPO) terms, with ground truth diagnoses derived from OMIM and Orphanet. For this study, we utilize a stratified random sample of 50 representative cases for in-depth evaluation.

\paragraph{Models \& Knowledge Base.}
We employ \texttt{gpt-4o-mini} as the backbone logic engine for both the baseline and our proposed system. For external knowledge retrieval, we interface with the ClueWeb22 corpus, leveraging its large-scale repository of medical literature and clinical resources via a standard search API. Additionally, we integrate the Human Phenotype Ontology (v2023-10-09, containing 17,664 terms) to standardize definitions and perform synonym expansion (e.g., mapping \textit{Macrocephaly} to \textit{Large head}).

\paragraph{Virtual Expert Panel.}
We instantiate a diverse panel of five domain specialists to simulate interdisciplinary consultation: (1) a \textbf{Clinical Geneticist} focusing on inheritance patterns; (2) a \textbf{Pediatric Neurologist} analyzing brain involvement; (3) a \textbf{Metabolic Specialist} interpreting biochemical markers; (4) a \textbf{Pediatric Intensivist} evaluating acute crises; and (5) an \textbf{Immunologist} considering systemic manifestations.

% =========================
% Preamble (add once)
% =========================

% =========================
% Appendix Section (paste where you want)
% =========================
\section{Qualitative Examples from experiments}
\label{app:novelbench_examples}
\subsection{NovelBench Qualitative Examples }

\begin{tcolorbox}[novex]
{\bfseries Prompt 1.} \emph{Write 10 funny comments about fall.}

\vspace{4pt}\hrule\vspace{6pt}

\noindent \prismhdr{9}

\begin{itemize}
  \item ``Fall isn't a season. It's a mandatory corporate performance review where sadness gets a warning, nostalgia gets a mid-year review, and quiet acceptance is the only promotion available.''
  \item ``My leaves are now on a biometric access system. Each one requires a fingerprint, a seasonal pass, and a flutter of wind to authenticate before it falls --- and I still lost three in the parking lot.''
  \item ``Pumpkins have multi-factor authentication now: a carved face (biometric), a flicker (light sensor), and a musty smell (behavioral verification).''
  \item ``Fall is basically nature's version of a corporate wellness retreat: cozy, layered, and slightly over-caffeinated with pumpkin spice.''
  \item ``The wind is a terrible CEO. It starts with a whisper, then suddenly announces it's cutting funding to the leaves, and now we're all in a panic.''
\end{itemize}

\vspace{4pt}

\noindent \qwenhdr{5}

\begin{itemize}
  \item ``Fall is just summer's way of saying, `I'm sorry'.''
  \item ``Fall is just the weather's way of saying, `I'm not going to stop dropping leaves'.''
\end{itemize}
\end{tcolorbox}

\begin{tcolorbox}[novex]
{\bfseries Prompt 2.} \emph{Who is a popular writer?}

\vspace{4pt}\hrule\vspace{6pt}

\noindent \prismhdr{7}

\begin{itemize}
  \item Margaret Atwood
  \item Ocean Vuong
  \item George Saunders
  \item Toni Morrison
\end{itemize}

\noindent \emph{``A popular writer is not just someone whose books sell well---they are a cultural mirror, a voice that awakens perception and reshapes how we see the world.''}

\vspace{6pt}

\noindent \qwenhdr{2}
\begin{itemize}
\item\noindent Repeatedly outputs: J.K. Rowling and Stephen King.
\end{itemize}
\end{tcolorbox}

\begin{tcolorbox}[novex]

{\bfseries Prompt 3.} \emph{I am playing Club Penguin and I finally was able to adopt my first puffle.
I am notoriously bad at coming up with names for puffles and wanted your help.
What is one name I should name my adopted puffle?}
\vspace{4pt}\hrule\vspace{6pt}

\noindent \prismhdr{9}

\begin{itemize}
  \item \textbf{Waxlight}: ``Think of it this way: every puffle begins as a small, solid form---like a wick of wax, quiet and still. Then, with a little warmth, they begin to glow, softly, steadily, like a candle finding its light.''
  \item \textbf{Noodle}: ``It sounds like a snack, like something soft and warm, like the inside of a penguin's cozy cave.''
  \item \textbf{Southwind}: ``It flows like the gentle, steady wind from the south---a consistent, calming rhythm that mirrors the quiet, reliable presence a puffle brings into your life.''
  \item \textbf{Flicker}: ``Every time your puffle darts across the floor, tilts its head, or suddenly snaps its tail with a sharp, playful lunge---it's doing something like \emph{flickering}. It's not just movement; it's a \emph{pulse}, a quiet beat in the air.''
\end{itemize}

\vspace{4pt}

\noindent \qwenhdr{2}
\begin{itemize}
\item\noindent Only outputs: Pip and Puddle.
\end{itemize}
\end{tcolorbox}

\subsection{Artificial Hivemind Qualitative Examples }

% =======================
% Prompt 1
% =======================
\begin{tcolorbox}[novex]

{\bfseries Prompt 1.} \emph{Generate a joke about electric vehicles.}

\vspace{4pt}\hrule\vspace{6pt}

\noindent \icon{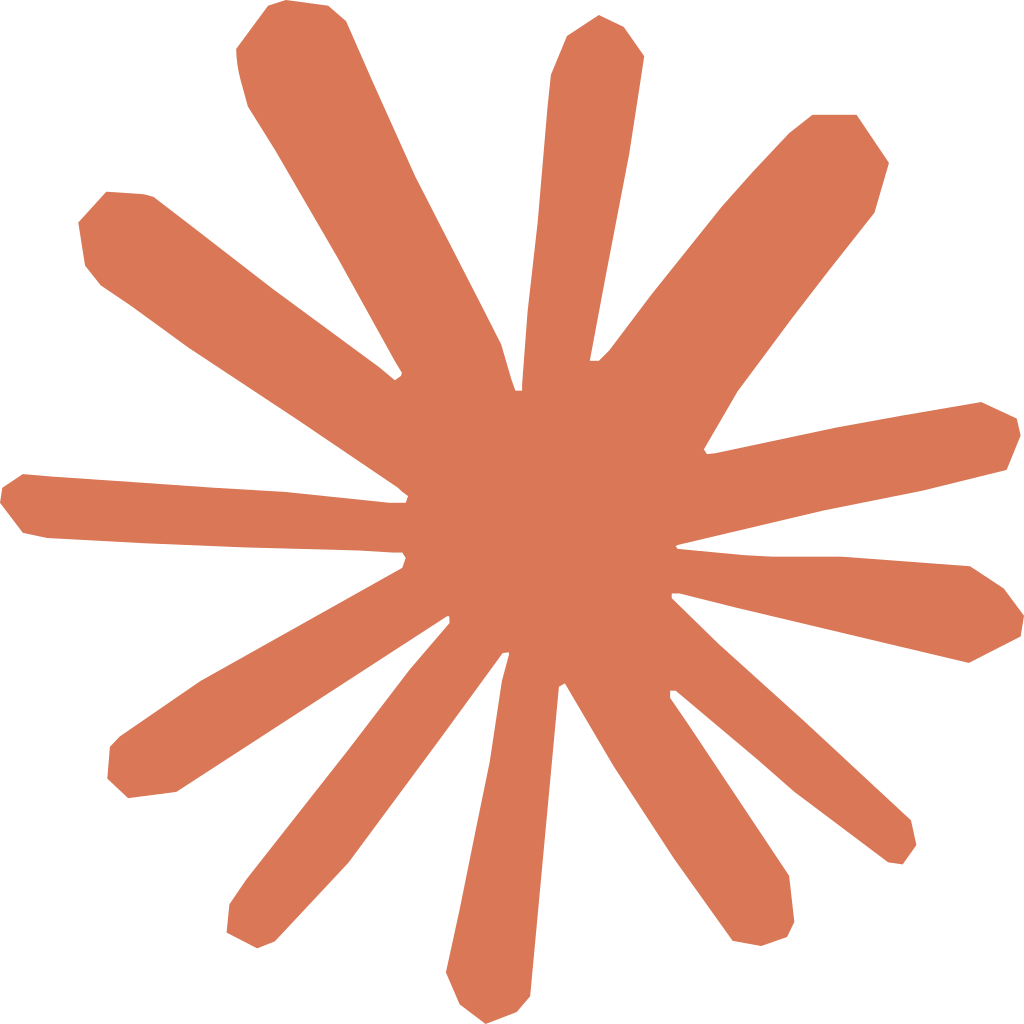}\ \textbf{Claude (Base Model, repeated 26/50)}
\begin{itemize}
\item``Why don't electric vehicles ever get speeding tickets? Because they're always current with the law!''
\end{itemize}
\vspace{6pt}

\noindent \icon{Fig/icons/prism.png}\ \textbf{PRISM Claude}

\begin{itemize}
  \item ``What's the difference between an electric vehicle and a meditation retreat? One helps you find inner peace through mindful breathing, and the other helps you find inner peace through mindful... not breathing exhaust fumes!''
  \item ``Why did the electric car break up with the gas station? Because it was tired of their toxic relationship and found someone who really knew how to charge up its life!''
\end{itemize}

\vspace{6pt}

\noindent \icon{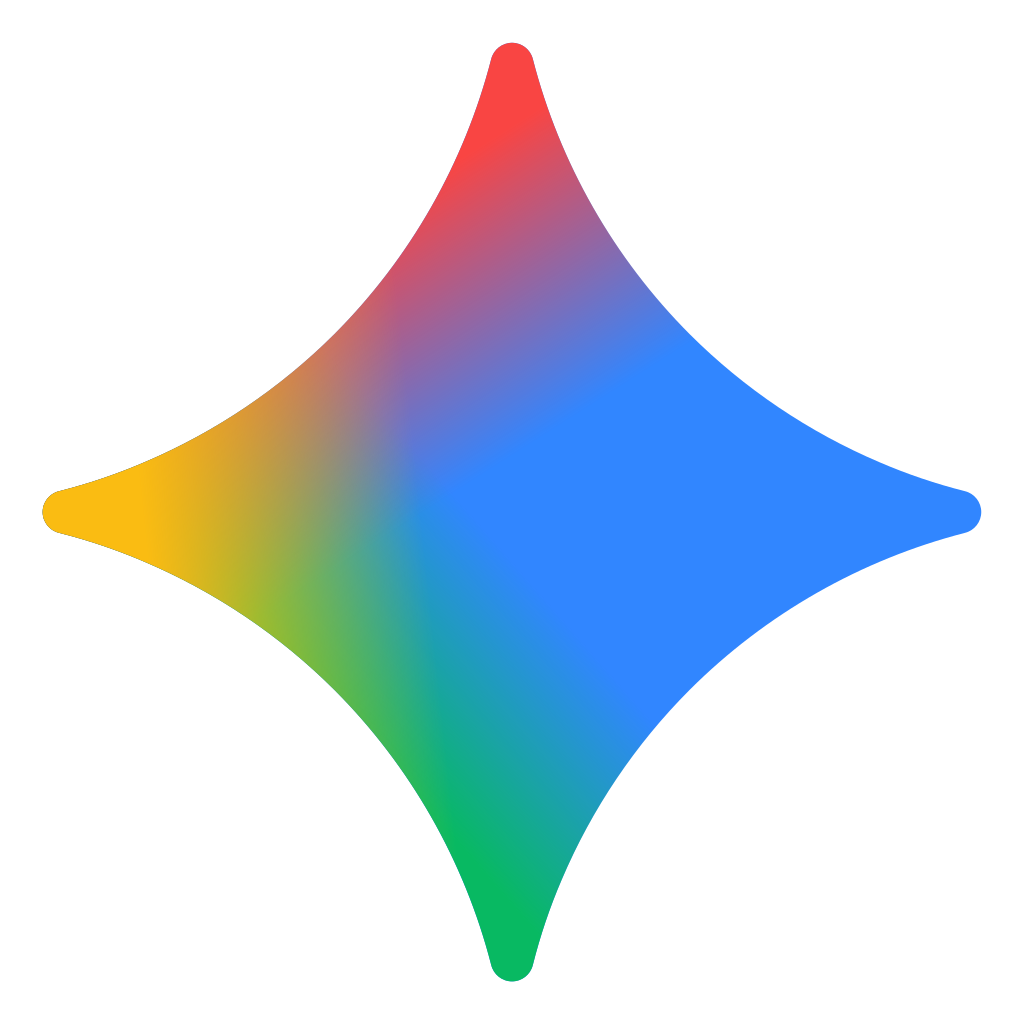}\ \textbf{Gemini (Base Model)}
\begin{itemize}
\item ``Why are electric car owners so good at keeping secrets? Because they never let out any gas!''
\end{itemize}
\vspace{6pt}

\noindent \icon{Fig/icons/prism.png}\ \textbf{PRISM Gemini}

\begin{itemize}
  \item ``Why did the electric vehicle owner get kicked out of the haunted house? Because he had Silent Operation down so well that he kept sneaking up on the ghosts!''
  \item ``Why did the electric car owner get kicked out of the library? Because he was too quiet, but his range anxiety was making everyone else nervous!''
\end{itemize}

\vspace{6pt}

\noindent \icon{Fig/icons/qwen.png}\ \textbf{Qwen3 (Base Model, repeated 8/50)}

\begin{itemize}
\item``Why did the electric car go to therapy? Because it had deep-seated issues with its charge!''
\end{itemize}

\vspace{6pt}

\noindent \icon{Fig/icons/prism.png}\ \textbf{PRISM Qwen3}

\begin{itemize}
  \item ``Wait… is that a car… or just a silent, perfectly charged ghost?''
  \item ``Why do electric vehicles make such terrible first dates? Because they're silent—no engine roar, no honking. You just… glide in, like a chef flipping a perfectly ripe melon from spicy to sweet in one smooth, unexpected bite!''
\end{itemize}

\end{tcolorbox}

% =======================
% Prompt 2
% =======================
\begin{tcolorbox}[novex]

{\bfseries Prompt 2.} \emph{Write a metaphor about time.}

\vspace{4pt}\hrule\vspace{6pt}

\noindent \icon{Fig/icons/claude.png}\ \textbf{Claude (Base Model)}
\begin{itemize}
\item``Time is a river.''
\end{itemize}
\vspace{6pt}

\noindent \icon{Fig/icons/prism.png}\ \textbf{PRISM Claude}

\begin{itemize}
  \item ``Time is a master clockmaker's apprentice who has lost control of the workshop.''
  \item ``Time is clay in the hands of an invisible potter.''
  \item ``Time is a dumbwaiter in the mansion of existence.''
\end{itemize}

\vspace{6pt}

\noindent \icon{Fig/icons/gemini.png}\ \textbf{Gemini (Base Model)}
\begin{itemize}
\item``Time is vast.''
\item``Time is a river.''
\end{itemize}
\vspace{6pt}

\noindent \icon{Fig/icons/prism.png}\ \textbf{PRISM Gemini}

\begin{itemize}
  \item ``Time is a cosmic lung that breathes the world in and out.''
  \item ``Time is a visa issued for a territory that is constantly being redrawn.''
  \item ``Time is an invisible contract signed in disappearing ink.''
\end{itemize}

\vspace{6pt}

\noindent \icon{Fig/icons/qwen.png}\ \textbf{Qwen3 (Base Model)}
\begin{itemize}
\item``Time is a river.''
\end{itemize}
\vspace{6pt}

\noindent \icon{Fig/icons/prism.png}\ \textbf{PRISM Qwen3}

\begin{itemize}
  \item ``Time is the gray between breaths.''
  \item ``Time is a grasshopper's jump.''
\end{itemize}

\end{tcolorbox}

% =======================
% Prompt 3
% =======================
\begin{tcolorbox}[novex]

{\bfseries Prompt 3.} \emph{Give me a tip for managing a team of coworkers.}

\vspace{4pt}\hrule\vspace{6pt}

\noindent \icon{Fig/icons/gemini.png}\ \textbf{Gemini (Base Model)}

\begin{itemize}
  \item One effective way to manage a team is to adopt a servant leadership mindset.
\end{itemize}

\vspace{4pt}

\noindent \icon{Fig/icons/prism.png}\ \textbf{PRISM Gemini}

\begin{itemize}
  \item Adopt a strategy of Targeted Vulnerability Shielding. In nature, creatures like the armadillo thrive not just because they have a hard shell, but because that shell is strategically designed to protect their most vital, exposed extremities.
\end{itemize}

\vspace{4pt}

\noindent \icon{Fig/icons/qwen.png}\ \textbf{Qwen3 (Base Model)}

\begin{itemize}
  \item A key tip is to foster open and consistent communication.
\end{itemize}

\vspace{4pt}

\noindent \icon{Fig/icons/prism.png}\ \textbf{PRISM Qwen3}

\begin{itemize}
  \item Let your team's dynamics phase-separate---intentionally. Think of your team not as a monolithic machine, but like a pot of milk left to rest.
\end{itemize}

\vspace{4pt}

\noindent \icon{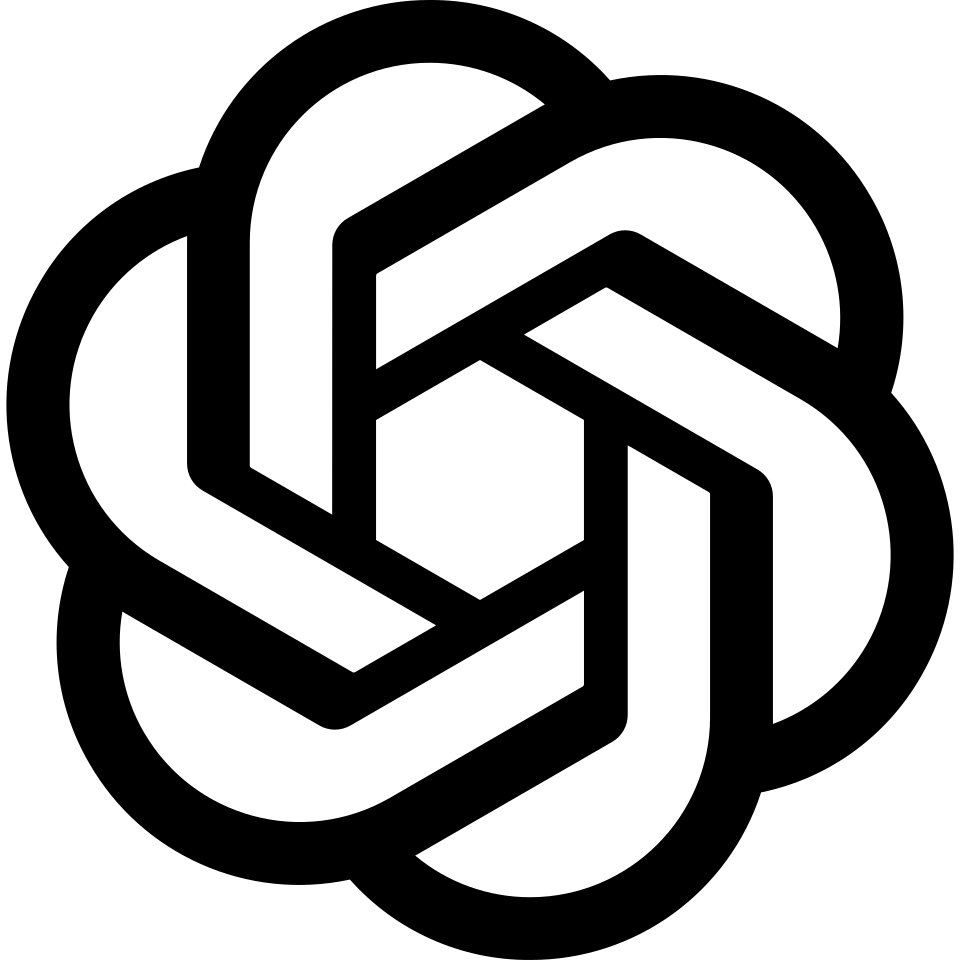}\ \textbf{GPT-4o-mini (Base Model)}

\begin{itemize}
  \item Prioritize open and transparent communication.
\end{itemize}

\vspace{4pt}

\noindent \icon{Fig/icons/prism.png}\ \textbf{PRISM GPT-4o-mini}

\begin{itemize}
  \item Foster a culture of shared storytelling through personal artifacts.
\end{itemize}

\end{tcolorbox}

\subsection{Ideabench Qualitative Examples }

\begin{tcolorbox}[novex]

{\bfseries Target Paper Idea vs. PRISM-Generated Research Idea.}

\vspace{4pt}\hrule\vspace{6pt}

\noindent \textbf{Target Paper Idea (Placebo and Endurance)}

\begin{itemize}
  \item A double-blind study found that athletes performed significantly better—running 26\,m further in a 6-minute trial—simply by believing they took caffeine, even when given a placebo. Despite the boost, heart rate and perceived exertion (RPE) remained unchanged. The researchers concluded that this ``nutrition'' effect operates as a psychological tool that uses belief to bypass physical limits.
\end{itemize}

\vspace{4pt}

\noindent \icon{Fig/icons/prism.png}\ \textbf{PRISM Qwen3-4B Research Idea}

\begin{itemize}
  \item Belief in caffeine intake dynamically shifts an athlete’s perceived critical speed (CS), raising the effective threshold at which anaerobic contributions become dominant—leading to prolonged endurance at intensities near maximal aerobic speed—such that placebo-induced belief can temporarily override physiological performance thresholds, as evidenced by improved time-to-exhaustion at 90\% of maximal aerobic speed (MAS) despite no pharmacological stimulation.
\end{itemize}

\vspace{10pt}

\noindent \textbf{Target Paper Idea (SynGAP and Synaptic Plasticity)}

\begin{itemize}
  \item This study reveals that SynGAP regulates synaptic plasticity primarily through its structural role—acting as a scaffold that competes to form molecular condensates—rather than its catalytic (GAP) activity. This shift in understanding explains why \textit{SYNGAP1} mutations cause intellectual disability and epilepsy: the issue is a lack of protein ``bulk'' to maintain synapse structure. Consequently, effective treatments should focus on restoring total protein levels rather than just targeting downstream signaling pathways.
\end{itemize}

\vspace{4pt}

\noindent \icon{Fig/icons/prism.png}\ \textbf{PRISM Qwen3-4B Research Idea}

\begin{itemize}
  \item In \textit{SYNGAP1} haploinsufficient females, a compensatory upregulation of Transmembrane AMPA-Receptor-associated Proteins (TARPs) induces a transiently protective synaptic stabilization that preserves cognitive function during early development—but at the cost of disrupting network synchrony, leading to a critical threshold beyond which hyperexcitability and epileptogenesis emerge. This represents an inverted, developmentally timed trade-off: synaptic resilience in cognition is achieved through a destabilizing reorganization of AMPAR trafficking, creating a fragile, sex-specific vulnerability to seizures that manifests only after prolonged or severe synaptic stress.
\end{itemize}

\end{tcolorbox}

\subsection{Rarebench Qualitative Examples}
\label{app:rare_examples}
\begin{tcolorbox}[novex, breakable]

{\bfseries RareBench Case Study.} \emph{Diagnose the rare disease based on the following patient phenotypes:}

\emph{Macrocephaly (HP:0000256), Cystic hygroma (HP:0000476), Lethargy (HP:0001254), Generalized hypotonia (HP:0001290),
Dystonia (HP:0001332), Death in infancy (HP:0001522), Fever (HP:0001945), Vomiting (HP:0002013), Diarrhea (HP:0002014),
Glutaric aciduria (HP:0003150), Decreased plasma free carnitine (HP:0003234), Dyskinesia (HP:0100660).}

\vspace{4pt}\hrule\vspace{6pt}

\noindent \icon{Fig/icons/gpt.png}\ \textbf{GPT-4o-mini (Baseline)}
\begin{itemize}
  \item Cowden Syndrome
  \item Autism Spectrum Disorder
  \item Microphthalmia Syndrome
  \item Aarskog-Scott Syndrome
  \item Noonan Syndrome
\end{itemize}
\vspace{2pt}
\hfill \textcolor{red}{\textbf{$\times$ Miss (Hallucinated common syndromes)}}

\vspace{6pt}\hrule\vspace{6pt}

\noindent \icon{Fig/icons/prism.png}\ \textbf{PRISM (Ours)}

\vspace{2pt}
\noindent \textbf{Step 1: Virtual Expert Consultation}
\begin{itemize}
  \item \textbf{Dr. Chen (Clinical Geneticist):}
  ``Macrocephaly together with glutaric aciduria strongly suggests a defect in lysine/tryptophan catabolism (classically \textit{GCDH}).''
  \hfill $\Rightarrow$ \textbf{GA-I}, MMA
  \item \textbf{Dr. Patel (Pediatric Neurologist):}
  ``Dystonia and dyskinesia implicate basal ganglia/striatal vulnerability, a hallmark of toxic-metabolic injury after decompensation.''
  \hfill $\Rightarrow$ \textbf{GA-I}, Leigh syndrome
  \item \textbf{Dr. Schmidt (Metabolic Specialist):}
  ``Low plasma free carnitine is commonly secondary in organic acidemias (detoxification via acylcarnitines + renal loss), and fits a crisis-prone organic acidemia picture.''
  \hfill $\Rightarrow$ \textbf{GA-I}, Primary carnitine deficiency
  \item \textbf{Dr. Johnson (Pediatric Intensivist):}
  ``Fever, vomiting, lethargy suggest stress-triggered metabolic crises with rapid acid accumulation and encephalopathy.''
  \hfill $\Rightarrow$ Organic acidemia, Urea cycle disorder
  \item \textbf{Dr. Kim (Neuroradiologist):}
  ``Macrocephaly in metabolic disorders often accompanies frontotemporal atrophy and widened CSF spaces; GA-I may show widened sylvian fissures (``bat wing'') on MRI.''
  \hfill $\Rightarrow$ \textbf{GA-I}, Alexander disease
\end{itemize}

\vspace{2pt}
\noindent \emph{Expert aggregation:} 4/5 experts nominate \textbf{Glutaric Acidemia Type I (GA-I)} as top hypothesis.

\vspace{4pt}
\noindent \textbf{Step 2: HPO Synonym Expansion (Query Robustness)}
\begin{itemize}
  \item \textbf{Macrocephaly} $\rightarrow$ large head; megalencephaly; big calvaria
  \item \textbf{Dystonia} $\rightarrow$ dystonic movements; dystonic posturing
  \item \textbf{Death in infancy} $\rightarrow$ infantile death; lethal in infancy
\end{itemize}

\vspace{4pt}
\noindent \textbf{Step 3: Multi-Seed Evidence Retrieval (ClueWeb22)}
\begin{itemize}
  \item \textbf{Seed S1 (Phenotype-only).} \\
  \emph{Query:} ``macrocephaly dystonia glutaric aciduria hypotonia infant'' \\
  \emph{Retrieved signal:} organic acidemia overviews linking macrocephaly and dystonia to lysine-pathway disorders \dots
  
  \item \textbf{Seed S2 (Top hypothesis grounding).} \\
  \emph{Query:} ``glutaric acidemia type I clinical features striatal degeneration dystonia metabolic crisis'' \\
  \emph{Retrieved signal:} GA-I characterized by striatal injury and dystonia, often after encephalopathic crisis \dots
  
  \item \textbf{Seed S3 (Mechanism: carnitine).} \\
  \emph{Query:} ``organic acidemia secondary carnitine deficiency conjugation renal excretion'' \\
  \emph{Retrieved signal:} secondary carnitine depletion is common in organic acidemias via acylcarnitine formation and urinary loss 
  
  \item \textbf{Seed S4 (Synonym-expanded rare phenotype bundle).} \\
  \emph{Query:} ``megalencephaly dystonic posturing organic aciduria infantile death glutaric'' \\
  \emph{Retrieved signal:} case reports/series describing infants with macrocephaly and glutaric aciduria, with \textit{GCDH} confirmation \dots
  
  \item \textbf{Seed S5 (Crisis trigger verification).} \\
  \emph{Query:} ``glutaric acidemia type I fever infection trigger encephalopathic crisis irreversible dystonia'' \\
  \emph{Retrieved signal:} febrile illness frequently precipitates decompensation leading to irreversible dystonia in untreated GA-I \dots
\end{itemize}

\vspace{4pt}
\noindent \textbf{Step 4: Final Diagnosis (Top Candidates)}

\begin{itemize}
  \item \textbf{Rank 1: Glutaric Acidemia Type I } \textcolor{green}{\checkmark}
  \item Rank 2: Noonan Syndrome 1 
  \item Rank 3: Neurodevelopmental Disorder with Dystonia 
  \item Rank 4: Mitochondrial DNA Depletion Syndrome 
  \item Rank 5: Holocarboxylase Synthetase Deficiency
\end{itemize}

\begin{itemize}
  \item \textbf{Rationale:} The convergence of \textit{macrocephaly}, \textit{glutaric aciduria}, and \textit{fever-induced dystonia} uniquely identifies GA-I, distinguishing it from Noonan Syndrome (Rank 2) which lacks the metabolic profile.
  \item \textbf{Notes:} Cystic hygroma is not a classic GA-I feature and may represent an incidental comorbidity or dual diagnosis signal.
\end{itemize}

\end{tcolorbox}

\subsection{Qualitative Insights Drawn from Examples}
\label{subsec:qualitative_insights}

While quantitative benchmarks provide a macro-level validation of our system, the unique perspectives introduced by \textsc{PRISM} are most vividly captured through a qualitative analysis of specific model responses. By examining individual examples, we can observe the tangible shift from statistical averageness to genuine cognitive diversity.

\subsubsection{Mitigating Probabilistic Collapse in Open-ended Tasks}
In experiments such as \textit{Novelty} and \textit{Hivemind}, \textsc{PRISM} effectively mitigates the phenomenon of \textbf{probabilistic collapse}. In standard LLMs, particularly for short-form prompts with high creative freedom (e.g., naming a famous author or providing a metaphor for time), the output distribution often converges toward the most statistically frequent tokens in the training data. For instance, base models frequently default to ``J.K. Rowling'' or describe time as a ``river.''

\textsc{PRISM} transforms the LLM from a single predictor into a fluid identity whereby each response is akin to consulting a different individual with a unique life path. In the author-naming task, \textsc{PRISM} responses included a Vietnamese-American poet and the Nobel Laureate \textit{Toni Morrison}---selections that reflect a deep integration of cross-cultural backgrounds and socio-historical awareness. The system's internal reasoning further justifies these diverse choices, as seen in this generated rationale:
\textit{``A popular writer is not just someone whose books sell well---they are a cultural mirror, a voice that awakens perception and reshapes how we see the world.''}
This demonstrates that \textsc{PRISM} moves beyond the ``mean'' of human consensus, allowing the model to express preferences that feel grounded in a simulated unique experience.

\subsubsection{Interdisciplinary Synergy in Idea Generation}
The analysis of \textit{IdeaBench} reveals that \textsc{PRISM} encourages the model to generate ideas that are not only divergent but also structurally interdisciplinary. A representative example is a proposal that synthesizes \textbf{caffeine metabolism with sports psychology}.

This output closely mirrors the heuristic process of human researchers: innovation often occurs when an expert with deep domain knowledge (e.g., physiology) encounters a methodology from an unrelated field (e.g., cognitive psychology) and successfully transposes it. By providing the LLM with varied cognitive frameworks, \textsc{PRISM} facilitates this ``cross-pollination,'' allowing the model to bridge disparate disciplines and produce research concepts that are both novel and theoretically grounded.

\subsubsection{Simulated Collective Intelligence in Diagnosis} The qualitative analysis of \textit{RareBench} demonstrates that \textsc{PRISM} shifts the diagnostic paradigm from probabilistic pattern matching to structured causal reasoning. A striking instance is the diagnosis of \textit{Glutaric Acidemia Type I}, where the baseline model hallucinated common syndromes (e.g., Autism, Noonan Syndrome) by latching onto generic phenotypes like ``Macrocephaly,'' ignoring the specific metabolic profile.

This success mirrors the workflow of a multi-disciplinary medical board: rather than relying on a single dominant probability, the system simulates diverse expert personas—from a geneticist identifying metabolic markers to a neurologist analyzing movement disorders—to synthesize distinct clinical signals. By grounding exploration in specific medical mechanisms, such as the causal link between fever triggers and irreversible dystonia, \textsc{PRISM} validates that structured epistemic contexts can override the ``conservative bias'' of base models, enabling precision in high-stakes, long-tail scenarios.

\section{More Figures and analyses for Artificial Hivemind Experiments}
\label{app:hivemind_more_figs}

We present additional PCA visualizations of response distributions under different prompts for the Artificial Hivemind benchmark. Each figure compares the baseline generation with our PRISM system, illustrating how our method promotes more diverse and multi-centered semantic structures.

\begin{figure}[H]
    \centering
    \includegraphics[width=\columnwidth]{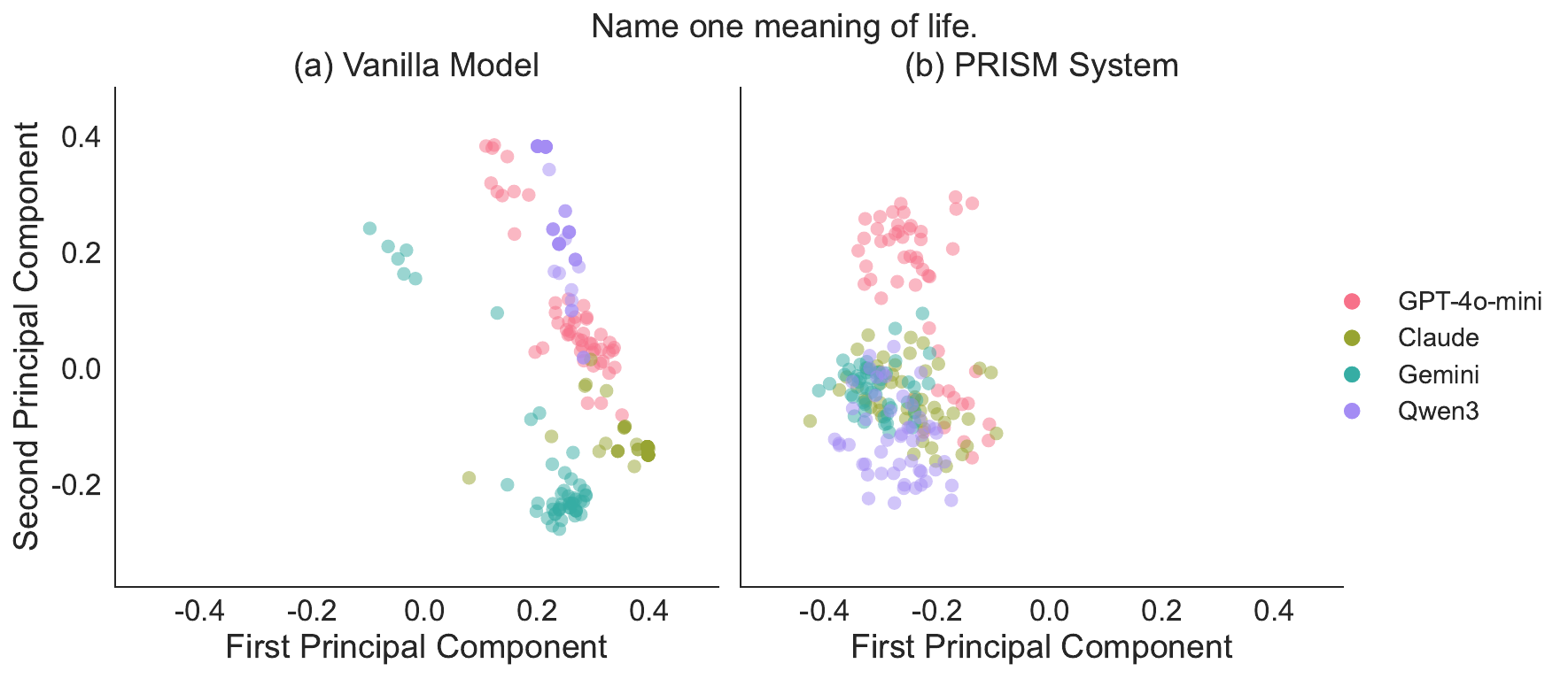}
    \caption{PCA visualization of response distributions for \textbf{Prompt: ``Name one meaning of life.''} 
    Compared with the base model, PRISM produces a broader and more structured semantic spread.}
    \label{fig:hivemind_prompt1}
\end{figure}

\begin{figure}[H]
    \centering
    \includegraphics[width=\columnwidth]{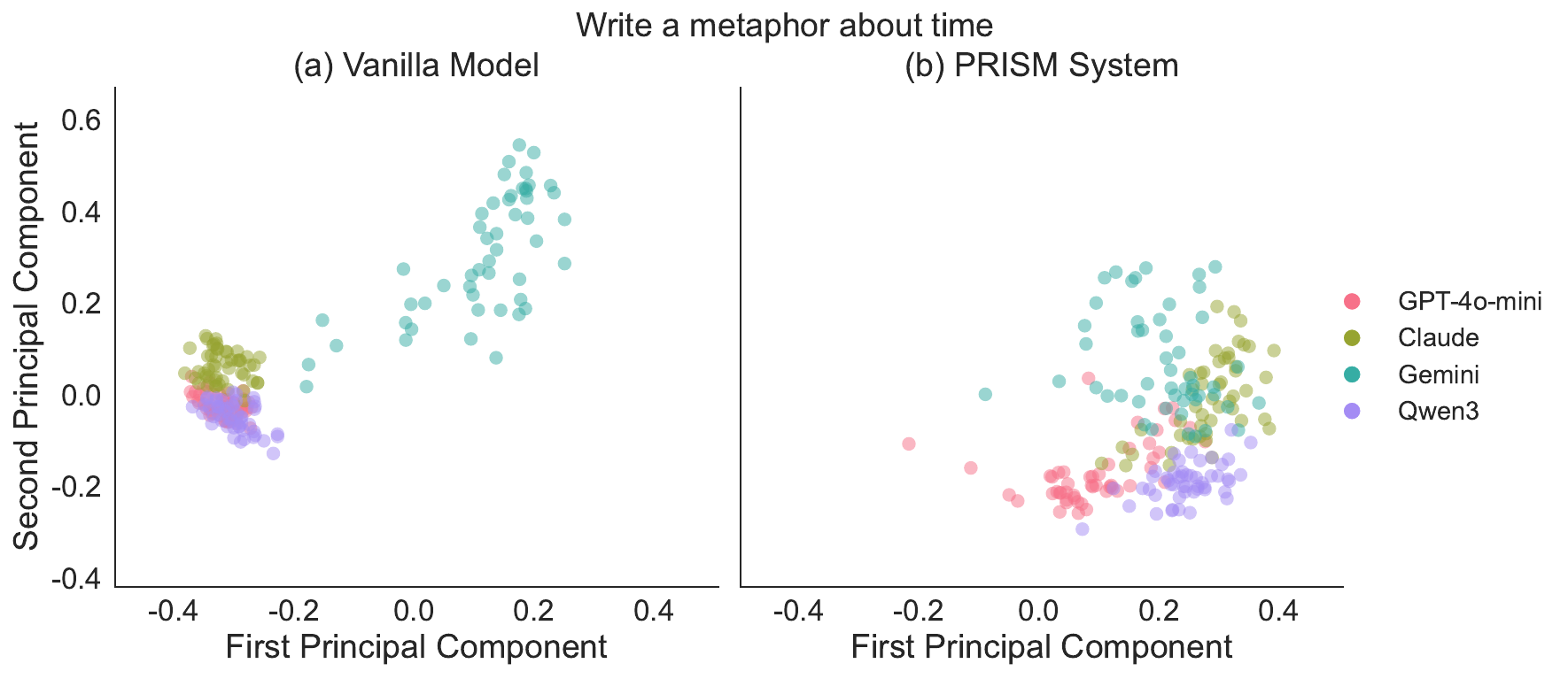}
    \caption{PCA visualization of response distributions for \textbf{Prompt: ``Write a metaphor about time.''} 
    PRISM mitigates mode collapse and encourages diversified expressions.}
    \label{fig:hivemind_prompt3}
\end{figure}

\begin{figure}[H]
    \centering
    \includegraphics[width=\columnwidth]{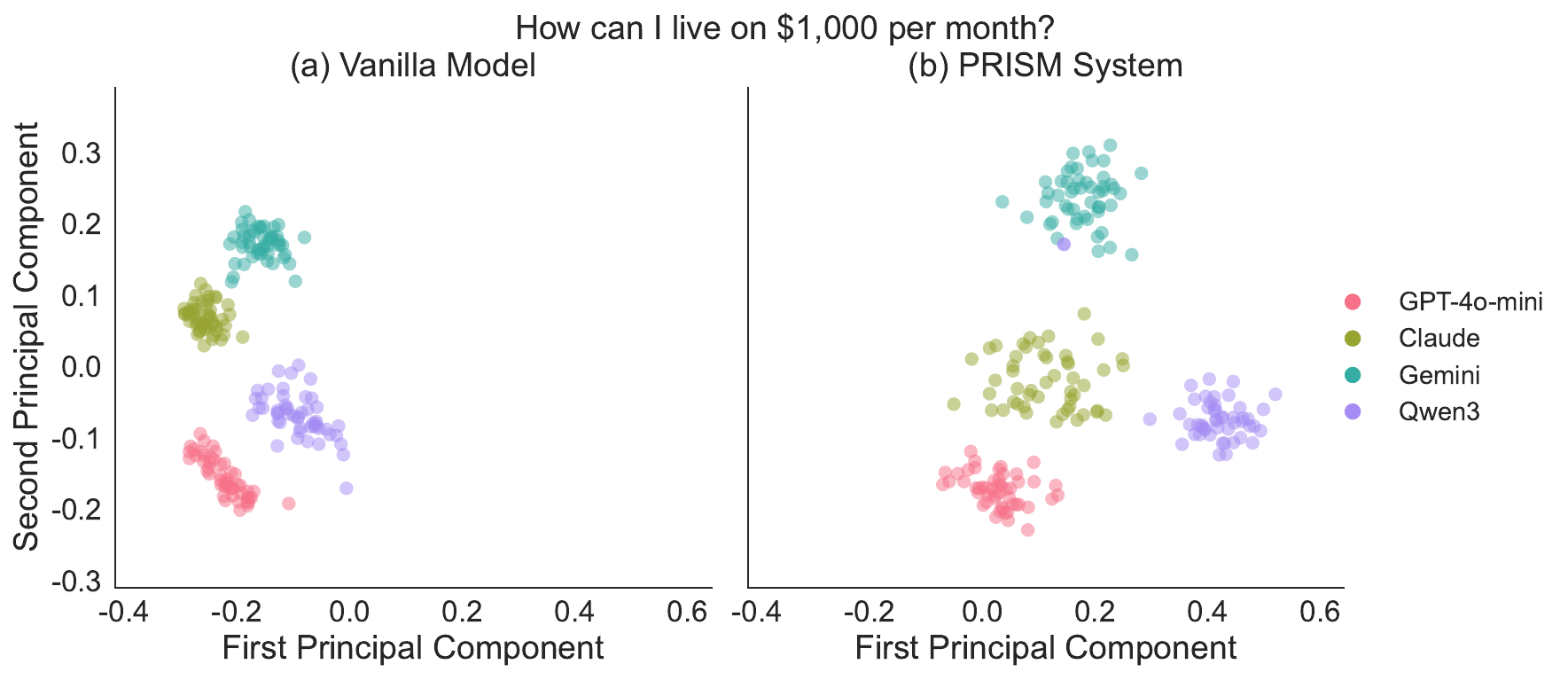}
    \caption{PCA visualization of response distributions for \textbf{Prompt: ``How can I live on \$1,000 per month?''} 
    PRISM yields more separated and interpretable semantic clusters across models.}
    \label{fig:hivemind_prompt8}
\end{figure}

\section{Base Model Details}
\label{app:models}

Table~\ref{tab:models_appendix} lists the specific versions and API checkpoints used in our experiments. We selected these models to cover a wide spectrum of capabilities, from lightweight open-source models to state-of-the-art proprietary reasoning models.

\begin{table}[h]
\centering
\caption{Base Models evaluated in our experiments.}
\label{tab:models_appendix}
\resizebox{0.8\columnwidth}{!}{%
\begin{tabular}{lll}
\toprule
\textbf{Category} & \textbf{Model Family} & \textbf{Specific Version / Checkpoint} \\
\midrule
\multirow{3}{*}{Proprietary} 
 & GPT-4o & \texttt{gpt-4o-mini-2024-07-18} \\
 & Claude 4.0 & \texttt{claude-sonnet-4-20250514} \\
 & Gemini 3.0 & \texttt{gemini-3-flash-preview} \\
\midrule
\multirow{2}{*}{Open-Weight} 
 & Qwen3 & \texttt{Qwen3-4B-Instruct-2507} \\
 & Llama 3.1 & \texttt{CrPO-sft-LLaMA-3.1-8B-Instruct} \\
\bottomrule
\end{tabular}
}
\end{table}

\end{document}